\documentclass{article}

\usepackage[preprint]{neurips_2025}

\usepackage[utf8]{inputenc} % allow utf-8 input
\usepackage[T1]{fontenc}    % use 8-bit T1 fonts
\usepackage{hyperref}       % hyperlinks
\usepackage{url}            % simple URL typesetting
\usepackage{booktabs}       % professional-quality tables
\usepackage{amsfonts}       % blackboard math symbols
\usepackage{nicefrac}       % compact symbols for 1/2, etc.
\usepackage{microtype}      % microtypography
\usepackage{xcolor}         % colors
\usepackage{amsmath}
\usepackage{graphicx}
\usepackage{multicol}
\usepackage{multirow}
\usepackage{makecell}
\usepackage{listings}

\title{Hi-Dyna Graph: Hierarchical Dynamic Scene Graph for Robotic Autonomy in Human-Centric Environments}

\author{Jiawei Hou, Xiangyang Xue${^*}$, Taiping Zeng
\thanks{Co-corresponding author. Jiawei Hou and Xiangyang Xue are with School of Computer Science, Fudan University, Shanghai, China{\tt\footnotesize \{jwhou23@m.fudan.edu.cn, xyxue@fudan.edu.cn\}}
Taiping Zeng is with Institute of Science and Technology for Brain-Inspired Intelligence, Fudan University, Shanghai, China{\tt\footnotesize \{zengtaiping@fudan.edu.cn\}}}}

\begin{document}

\maketitle

\begin{figure}[h]
\centering
\vspace{-0.4 in}
\includegraphics[width=\linewidth]{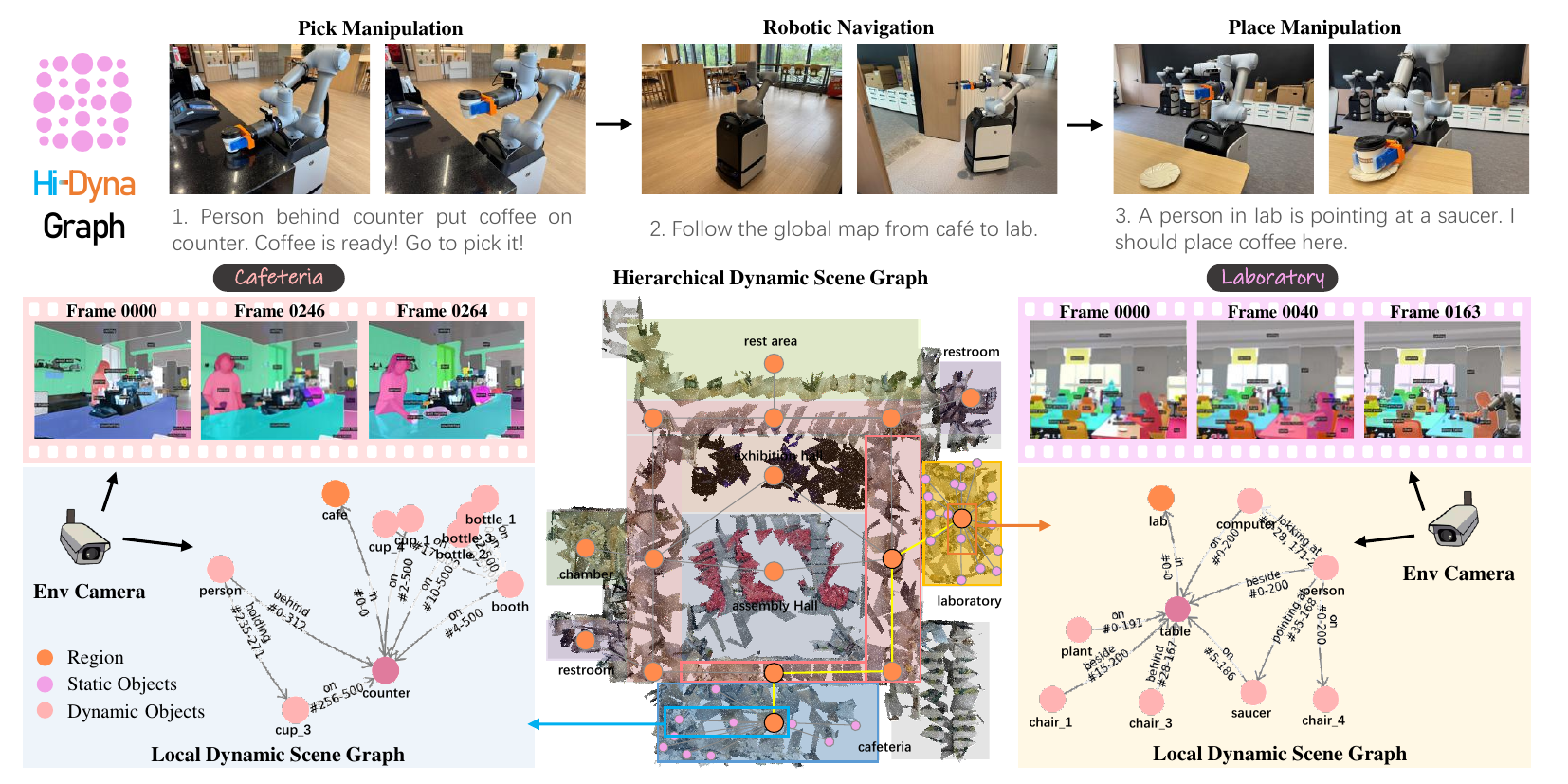}
\vspace{-0.25 in}
\caption{Hi-Dyna Graph creates a hybrid graph of global static layouts and local dynamic relations. Integrating such graphs into reasoning, robotics can manage autonomy in dynamic large scenes.}
\label{fig::teaser}
\end{figure}

\begin{abstract}
  % Autonomous operation of service robotics in human-centric scenes, such as home environments, remains challenging due to the need for real-time understanding of evolving dynamics and adaptive context-aware embodied reasoning. 
  Autonomous operation of service robotics in human-centric scenes remains challenging due to the need for understanding of changing environments and context-aware decision-making.
  While existing approaches like topological maps offer efficient spatial priors, they fail to model transient object relationships, whereas dense neural representations (e.g., NeRF) incur prohibitive computational costs. 
  Inspired by the hierarchical scene representation and video scene graph generation works, we propose Hi-Dyna Graph, a hierarchical dynamic scene graph architecture that integrates persistent global layouts with localized dynamic semantics for embodied robotic autonomy.
  Our framework constructs a global topological graph from posed RGB-D inputs, encoding room-scale connectivity and large static objects (e.g., furniture), while environmental and egocentric cameras populate dynamic subgraphs with object position relations and human-object interaction patterns. A hybrid architecture is conducted by anchoring these subgraphs to the global topology using semantic and spatial constraints, enabling seamless updates as the environment evolves. An agent powered by large language models (LLMs) is employed to interpret the unified graph, infer latent task triggers, and generate executable instructions grounded in robotic affordances.
  We conduct complex experiments to demonstrate Hi-Dyna Graph’s superior scene representation effectiveness. Real-world deployments validate the system’s practicality with a mobile manipulator: robotics autonomously complete complex tasks with no further training or complex rewarding in a dynamic scene as cafeteria assistant. See \url{https://anonymous.4open.science/r/Hi-Dyna-Graph-B326} for video demonstration and more details.
\end{abstract}

\section{Introduction}
Recent advancements in embodied intelligence have enabled robotics to interact with complex environments~\cite{werby23hovsg, Maggio2024Clio, hou2025topofield}, yet employing them autonomously working in human-centric dynamic scenes remains challenging. A critical barrier lies in enabling robotics to (1) efficiently manage multimodal scene information, (2) reason about ongoing activities in rapidly changing environments, and (3) autonomously generate and execute tasks based on evolving environmental changes and situational awareness. Robotics struggle with the unpredictability of human-centric environments where object states, spatial relationships, and task requirements shift continuously. These limitations stem from fundamental gaps in scene-understanding architectures that fail to unify persistent environmental knowledge with perceptual updates.

To conduct scene understanding, researchers have explored topological maps as memory-efficient and easily-queriable structural priors~\cite{blochliger2018topomap, gomez2020hybrid, topo1, topo2, topo3}. However, such approaches exhibit critical shortcomings: Static graph nodes cannot model transient object relationships (e.g., utensils moved during cooking), while rigid hierarchies collapse under concurrent updates to entities with varying dynamism (static furniture vs. frequently manipulated items). This creates a representational mismatch between the robotics internal world model and the actual environment state, particularly in zones of high human activity. On the other hand, dense scene representations such as Neural Radiance Field (NeRF)~\cite{nerf} and Gaussian Splattings (GS)~\cite{kerbl3Dgaussians} series approaches have explored introducing temporal embeddings and editing the representations to manage the dynamics~\cite{Wu20244DGS, attal2023hyperreel, hexplane, kplane}. However, most of these dense representations are computationally intensive and their dense volumetric nature hinders efficient querying for downstream tasks~\cite{hou2025topofield}. 

Recent works turn to hierarchically conduct the scene understanding which is an inspiring solution~\cite{hou2025topofield, werby23hovsg, Maggio2024Clio}. Topo-Field~\cite{hou2025topofield} leverages sparse topological map to represent the scene layouts and dense neural field for content details. Although this pipeline achieves efficiency in down-stream tasks while remains enough semantics and geometries, it can not manage dynamics based on the static scene assumption. Scene graph generation methods partially mitigate this by inferring object relationships from monocular observations~\cite{yang2023pvsg}. PSG4D~\cite{yang2023psg4d}, as an example, has demonstrated the ability to generate scene relation graph and help robotics reasoning. Yet these frameworks remain myopic because they lack mechanisms to maintain global spatial context, historical state tracking, or embodied agent perspectives. without integrating global scene context—limits applicability, it is hard for robotics to execute long-horizon tasks in large-scale environments for real-world demonstration.

Therefore, we aim to synergistically combine dynamic relation reasoning and hierarchical scene graphs. The hierarchical scene representation is separated into static and dynamic components based on spatiotemporal attributes. As shown in Fig. \ref{fig::teaser}, the static scene graph encodes rarely changing elements (e.g. architectural layouts, large furniture), while dynamic subgraphs maintain transient objects (e.g., small items, humans) and their evolving relationships. Dynamic subgraphs extend the temporal expressiveness of hierarchical representations, while the global scene graph provides spatial priors to expand the perceptual horizon of localized relation inference. This dual enhancement equips robotics with robust autonomy in dynamically evolving scenes.

Specifically, this work proposes Hi-Dyna Graph, a hierarchical dynamic scene graph which helps robotics understand and reason about the complex human-centric environments for self-driven embodied autonomy. From posed RGB-D inputs, we construct a global topological graph capturing layout-level semantics and huge objects that are rarely moved, while localized video streams from embodied cameras or environmental cameras populate dynamic subgraphs with object affordances and human interaction states. The dynamically graphs are anchored to global graph according to the semantic and spatial constraints. To bring this hybrid scene graph into robotic deployment, we integrate large language models (LLMs) as reasoners that interpret the unified graph, infer latent task triggers (e.g., unwashed dishes $\to$ initiate cleanup), generate executable instructions, and adapt plans as the graph evolves.
The conducted Hi-Dyna Graph is validated through complex experiments to show its superior effectiveness. Further, real-world robotic deployments are conducted based on Hi-Dyna Graph for embodied autonomy demonstration. 

Our key contributions can be concluded as: 
\begin{itemize}
    \item We propose Hi-Dyna Graph, a hierarchical dynamic scene graph for human-centric scene understanding, which introduces global scene graph as spatial prior and local relation graph as temporal adaptation. This dual enhancement equips robotics with embodied autonomy in dynamically evolving scenes.
    \item A hybrid scene graph architecture is conducted by anchoring dynamic subgraphs to the global topology with spatial and semantic constraints. An LLM-powered agent is employed to interpret, reason, and utilize this graph for embodied tasks grounded in robotic affordances. 
    \item We leverage real-world robotic deployments to demonstrate the effectiveness of our approach. Complex analyses are performed to show the superior performance of our strategy.

\end{itemize}

\section{Related Works}
\subsection{Dynamic Scene Representation}

Representing dynamic scenes has been an essential challenging extension for scene representation. Works like T-NeRF~\cite{tnerf,li2021neural,li2022neural,du2021neural,park2021hypernerf,park2021nerfies} extended NeRF~\cite{nerf} with additional time dimension or latent code. Gaussian Splatting (GS) series, as an explicit approach, also tried to adapt to dynamics~\cite{kerbl3Dgaussians,Wu20244DGS,yang2023deformable3dgs,spacegs}. 
% Deformable-3DGS~\cite{yang2023deformable3dgs} intrduced a deformation network to model the motion, while Spacetime-GS~\cite{spacegs} applied tracking on each Gaussian. 4DGS~\cite{Wu20244DGS} further optimize the network with a more compact one, resulting higher efficiency. 
However, dense representations are often computationally intensive and face challenges on efficient querying for downstream tasks~\cite{hou2025topofield}.
Unlike detailed scene reconstruction, topology-based representations address efficiency by abstracting environments into sparse graphs. Recent hierarchical representation approaches, such as HOV-SG~\cite{werby23hovsg} and Topo-Field~\cite{hou2025topofield}, introduce object-level embeddings with abstract topology to form hybrid representations. Yet based on static environments assumption, static graph vertices cannot model transient object relationships while rigid hierarchies collapse under concurrent updates to entities with varying dynamism. 
Our work bridges this gap through a hierarchical dynamic scene graph (Hi-Dyna Graph) that couples a persistent global topological map (encoding room layouts and static macro-objects) with dynamically updated local subgraphs based on video relation prediction. This dual representation preserves efficiency for large-scale navigation while maintaining granular, updatable semantics for task-oriented reasoning.

\subsection{Relation Graph Generation}
Significant progress has been made in inferring relation graphs from monocular video streams~\cite{yang2023pvsg,yang2023psg4d}. State-of-the-art methods combine panoptic segmentation with instance tracking to detect objects and predict inter-object relationships across frames. However, these approaches remain fundamentally myopic: Their reliance on single-view inputs limits awareness of occluded regions and global spatial context, artificially constraining a robot’s operational scope. For example, a robotic might recognize a "book on a desk" in its immediate view but remain oblivious to the desk’s location relative to the broader home layout without scene layout knowledge. Furthermore, existing relation graph generation frameworks operate as passive observers rather than embodied agents, they lack integration with robotic action loops and have not been validated in physical task execution. To overcome these limitations, our proposed architecture fuses localized dynamic subgraphs (derived from egocentric or environmental camera streams) with a global static scene graph. By grounding LLM-based task reasoning in this unified representation, our system not only interprets transient object relationships but also leverages persistent spatial knowledge to guide robotics through long-horizon activities.

\section{Overview}

\begin{figure}[t]
\centering
\includegraphics[width=\linewidth]{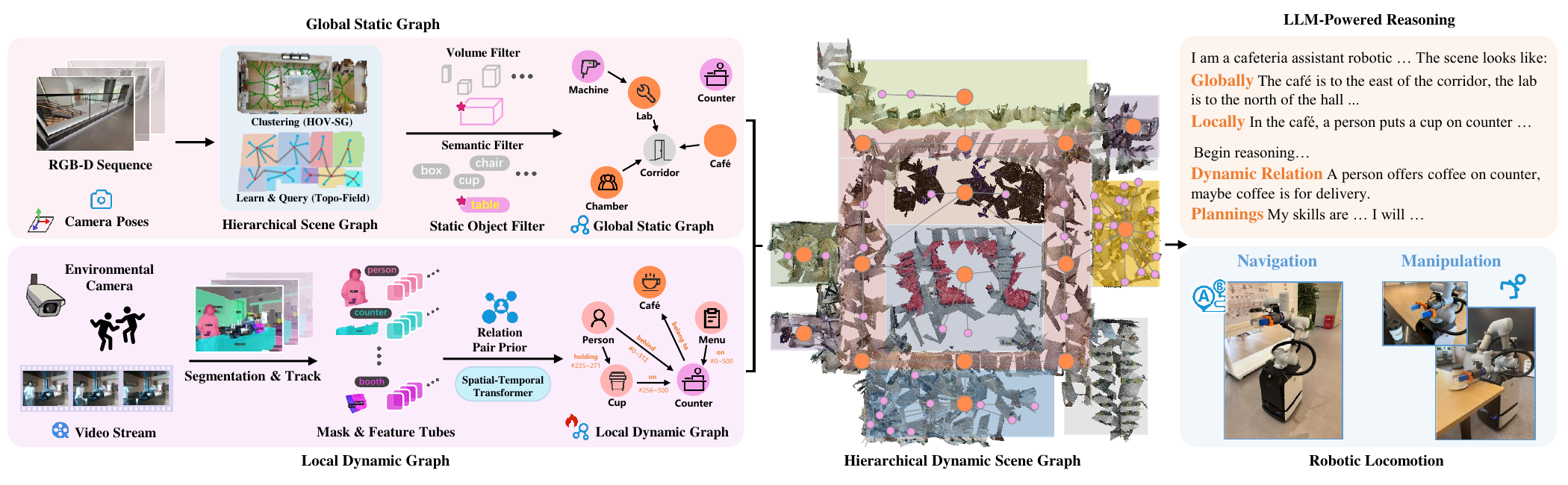}
\vspace{-0.25 in}
\caption{\textbf{Pipeline of our proposed Hi-Dyna Graph.} The hierarchical dynamic scene graph construction process consists of two branches, including the global static graph and local dynamic graph generation. The separate parts are combined to form a united scene graph representation. By employing LLM as reasoning approach, the scene graph is fed as prompts, together with other context, to drive the robotic mobile manipulator to manage task sequences.}
\vspace{-0.1 in}
\label{fig::method}
\end{figure}

This paper proposes Hi-Dyna Graph, a hierarchical dynamic scene graph to achieve embodied autonomy in dynamic environments. The scene graph can be noted as $\mathcal{G} = \{\mathcal{G}_s, \mathcal{G}_d\}$. Specifically,
\begin{equation}
\begin{aligned}
    \mathcal{G}_s: (\mathcal{V}_s, \mathcal{E}_s) &= \text{F}(\{I_k, T_k\}^N_{k=1}), \\
    \mathcal{G}_d: (\mathcal{V}_d, \mathcal{E}_d) &= \text{G}(\mathcal{F}_t).
\end{aligned}
\end{equation}

$\mathcal{G}_s$ captures persistent environmental layouts and macro-objects which is built from posed RGB-D images $\{I_k, T_k\}^N_{k=1} (T_k \in SE(3))$, and $\mathcal{G}_d$ is the dynamic local graph incrementally built from video streams $\mathcal{F}_t$. $(\mathcal{V}, \mathcal{E})$ represents the vertices and edges in the graph. $\text{F}$ and $\text{G}$ denote global static scene graph and local dynamic scene graph construction process.

Given posed RGB-D images of the environment, we encode the RGB images into vision-language embeddings and rise them to 3D space according to the depth and camera pose. We acquire the regions and macro-objects by querying this embedding point cloud and form the graph vertices $\mathcal{V}_s = (\mathcal{V}_r, \mathcal{V}_o)$, where $\mathcal{V}_r$ is the region vertice and $\mathcal{V}_o$ is the macro-object vertice. As for the dynamic scene graph, the process can be denoted as 
\begin{equation}
    \text{Pr}(\mathcal{V}_d, \mathcal{E}_d\ |\ \mathcal{F}_t) = \text{Pr} (M_t,O_t,R_t\ |\ \mathcal{F}_t), 
\end{equation}
where $M_t$ is the binary object mask tube, $O_t$ is the object label, and $R_t$ is the inter-object relation. The dynamic subgraph is anchored to the global graph according to semantic and spatial relations.

The scene graph serves as a structured knowledge base for LLM-based reasoning. By querying the hierarchical layers, LLM grounds with spatial and temporal context: the static graph provides global navigational constraints, while dynamic subgraphs supply localized task triggers. The LLM parses this multimodal input through prompt templates as
\begin{equation}
\label{equation::prompt}
    \underbrace{\text{You are}\dots}_{\text{system context}}\ 
    \underbrace{\text{Scene structures: }
    \mathcal{V}_s,\ \mathcal{E}_s}_{\text{structure from static graph }}\ 
    \underbrace{\text{Ongoing relations: }
    \mathcal{V}_d,\ \mathcal{E}_d}_{\text{activities from dynamic graph}}\ 
    \underbrace{\text{Optional skills: }\dots}_{\text{embodied primitives}}\ ,
\end{equation}
that integrate scene knowledge, skill primitives, and instructions, generating executable tasks.
Eventually, they are translated into mobile base navigation and robotic arm pick-place sequences.

\section{Method}
Our framework operates through four core modules: global static scene graph construction, local dynamic relation graph generation, hierarchical graph fusion, and LLM-driven task reasoning. The workflow is illustrated in Fig. \ref{fig::method}, with algorithmic details described bellow.

\subsection{Global Static Scene Graph}
\label{method::global_graph}
The global static scene graph $\mathcal{G}_s$ captures persistent environmental semantics and represents the layout structures through vertices $\mathcal{V}_s$ and edges $\mathcal{E}_s$.

\textbf{Topology Construction.} Given posed RGB-D images $\{I_k, T_k\}^N_{k=1} (T_k \in SE(3))$, we first extract the open-vocabulary embeddings with vision-language encoder from RGB images. These embeddings are rised up to 3D space by back-projection and translation according to the depth and camera pose. The 3D point-wise embeddings could be leveraged to form a topological map either by learning a neural field and querying objects and regions separately (e.g. Topo-Field~\cite{hou2025topofield}) or by clustering embeddings and employing polices for region segmentation (e.g. HOV-SG~\cite{werby23hovsg}).

\textbf{Relatively Static Objects Filter.} However, current scene graph construction approaches operate under a static scene assumption, resulting in graphs that capture observational snapshots at discrete moments. Such representations inevitably suffer from transient noise and lack generalizability, including false detections and dynamic objects frequently interacting with humans. In contrast, we adopt a relatively static assumption for scene modeling, positing that scene layouts and objects exceeding specific volume thresholds (e.g., large furniture) or belonging to designated semantic categories (e.g., couch, fridge, TV) tend to remain stationary, while other entities exhibit higher dynamism. Consequently, during static scene graph construction, we selectively establish vertices only for objects with bounding-box volumes surpassing threshold $\text{v}_{thr}$ or belonging to semantic class $\text{C}_s$, effectively filtering out transient or unstable elements. After this step, a hierarchical static topological graph is built with layouts and relatively static objects:
\begin{equation}
    \mathcal{G}_s:(\mathcal{V}_s,\mathcal{E}_s \ | \  \mathcal{V}_s = \{\mathbf{v}_{s}^r \cup \mathbf{v}_{s}^o\}, \mathcal{E}_s = \{\mathbf{e}_s^c \cup \mathbf{e}_s^b\}),
\end{equation}
where $\mathcal{V}_s$ consists of region vertices $\mathbf{v}_{s}^r$ and static object vertices $\mathbf{v}_{s}^o$. $\mathcal{E}_s$ consists of region connectivity relation edges $\mathbf{e}_s^c$ and object-region belonging relation edges $\mathbf{e}_s^b$.

\subsection{Local Dynamic Relation Graph}
\label{method::local_graph}
Local dynamic subgraphs $\mathcal{G}_d$ are built and updated incrementally from video streams $\mathcal{F}_t$. The video could come from an environmental camera or an embodied one whose global pose is available.

\textbf{Video Perception and Tracking.} For each video frame $F_i\in \mathbb{R}^{H\times W\times 3}$, we employ FC-CLIP~\cite{yu2023fcclip} for open-vocabulary segmentation which segments the image into a set of masks with associated features and semantic labels $\{(m_i,f_i,c_i)\}_{i=1}^K$, where $m_i$ is the binary mask, $f_i$ is the related semantic feature of the specific instance, $c_i$ is the class label. To keep align with the on-going activities in the environment, we further adapt a sliding window to continuously track the instances in each period with a time span of $\Delta t$. Specifically, inspired by PVSG~\cite{yang2023pvsg}, we employ UniTrack~\cite{wang2021unitrack} to link frames in each period and consequently obtain $N$ tracked instances.

\textbf{Relation Prediction.} Inspired by the Pair-Net~\cite{wang2024pair}, we predict the relations of instances in a subject-object pair form. Specifically, as described in PVSG~\cite{yang2023pvsg}. In contrast to conventional approaches, we incorporate our relatively static scene assumption when selecting Top-k relational pairs for prediction. We posit that humans in the scene are more likely to act as subjects, while large furniture items predominantly serve as objects. Consequently, we assign higher priority to relational pairs where humans act as subjects or large furniture serves as objects. Specifically, we allocate 70\% of the Top-k pairs from prioritized categories with the remaining 30\% drawn from others. Then we perform relation category prediction for each selected pair. Predicted relations in $\Delta t$ period can be denoted as 
\begin{equation}
\mathcal{R}_{\Delta t} = \{\text{id}_{sub}, \text{class}_{sub}, \text{id}_{obj}, \text{class}_{obj}, (t_a^l, t_b^l)_{l=1}^L\}_{j=1}^J,
\end{equation}
where id denotes the subject / object index of instances in such period and class is the semantic label. $(t_a^l, t_b^l)_{l=1}^L$ denotes the $L$ time spans that the relation is found. In each span, the relation happens at $t_a^l$ and ends at $t_b^l$. To enhance the temporal stability of predictions and mitigate fluctuations across time intervals, we consolidate consecutive temporal segments when the interval between the end of a preceding segment $t_b^{l-1}$ and the start of the subsequent segment $t_a^l$ is less than 2 seconds.

\subsection{Hierarchical Graph Fusion}
\label{method::graph_fusion}
To empower the dynamic graphs with global layouts and introduce dynamics into global graph, we anchor local dynamic graphs $\mathcal{G}_d$ to global static graph $\mathcal{G}_s$. Depending on whether the accurate camera pose and depth is available, we provide two approaches.

\textbf{Spatial Alignments.} If the camera pose and depth is available for the environmental or embodied camera, we simply back-project the instance vertices in the dynamic graph to 3D space and merge the vertices $\mathcal{V}_s$ and $\mathcal{V}_d$ from the global and local graph whose bounding-box overlap exceeds the threshold $b_{thr}$ (60\% in our experiments). For vertices in $\mathcal{V}_d$ that are not in the global graph, edges are added between them and the global region vertice, indicating the instances belong to the region.

\textbf{Semantic Matching.} If we only know the region that the camera belonging to instead of accurate camera pose and depth, we select the dynamic graph vertices $\mathcal{V}_d'$ whose predicted class belong to the designated semantic categories $\text{C}_s$, as mentioned in Section \ref{method::global_graph}, which tends to be the relatively static stuff. We merge the connected subgraphs $\mathcal{G}_d'$ from $\mathcal{G}_d$ that $\mathcal{V}_d'$ belongs to into the global graph $\mathcal{G}_s$ and remain the related vertices and edges unchanged. For other connected subgraphs, we add an edge between the region vertice and the subgraph indicating the belonging relationship.

At each $\Delta t$ period, the dynamic components in the graph are cleared and updated as described in Section \ref{method::local_graph} and \ref{method::graph_fusion} to keep align with the current environment. 

\subsection{LLM-Driven Task Reasoning and Execution}
To show how Hi-Dyna Graph can help robotics manage autonomy tasks, we introduce an LLM agent and parse the hierarchical dynamic scene graph into textual prompts for reasoning. The LLM generates task sequences through chain-of-thought prompting and manage the tasks with navigation and object pose estimation. The pipeline is constructed as follows.

\textbf{Chain of Thoughts.} As illustrated in Equation \ref{equation::prompt}, the prompts mainly consist of 1) a system instruction that describes the agent role, environment contexts, and the brief autonomy policy 2) text-formatted multilevel dynamic scene graph which is fed to the LLM at a constant frequency 3) optional skills that the robotics can manage according to the embodied ability. In each query, LLM is expected to describe the activities in the environment and conduct reasoning on whether optional skills can help with the activities. If any helpful action is available, a sequence of navigation and object pick-place tasks will be generated in order. The task instruction is formed as ``navigate to / pick / place \{object\} in \{region\}'' An example of prompt of a cafe assistant robotics is shown in the appendix.

\textbf{Navigation and Manipulation.} The navigation process follows a two stage planning strategy as described in ELA-ZSON~\cite{hou2025elazson}, where the robotic approach the target instance by querying the object veretices position and planning on the graph. After approaching the target via navigation, the embodied camera takes an image and query the target object to estimate the 6 degree-of-freedom (DoF) pose. Robotic arm takes this pose as input to conduct pose-guided manipulation. The detailed manipulation process can be referenced to Polaris~\cite{wang2024polaris}.

\section{Experimental Results}

\begin{figure}[t]
\centering
\includegraphics[width=\linewidth]{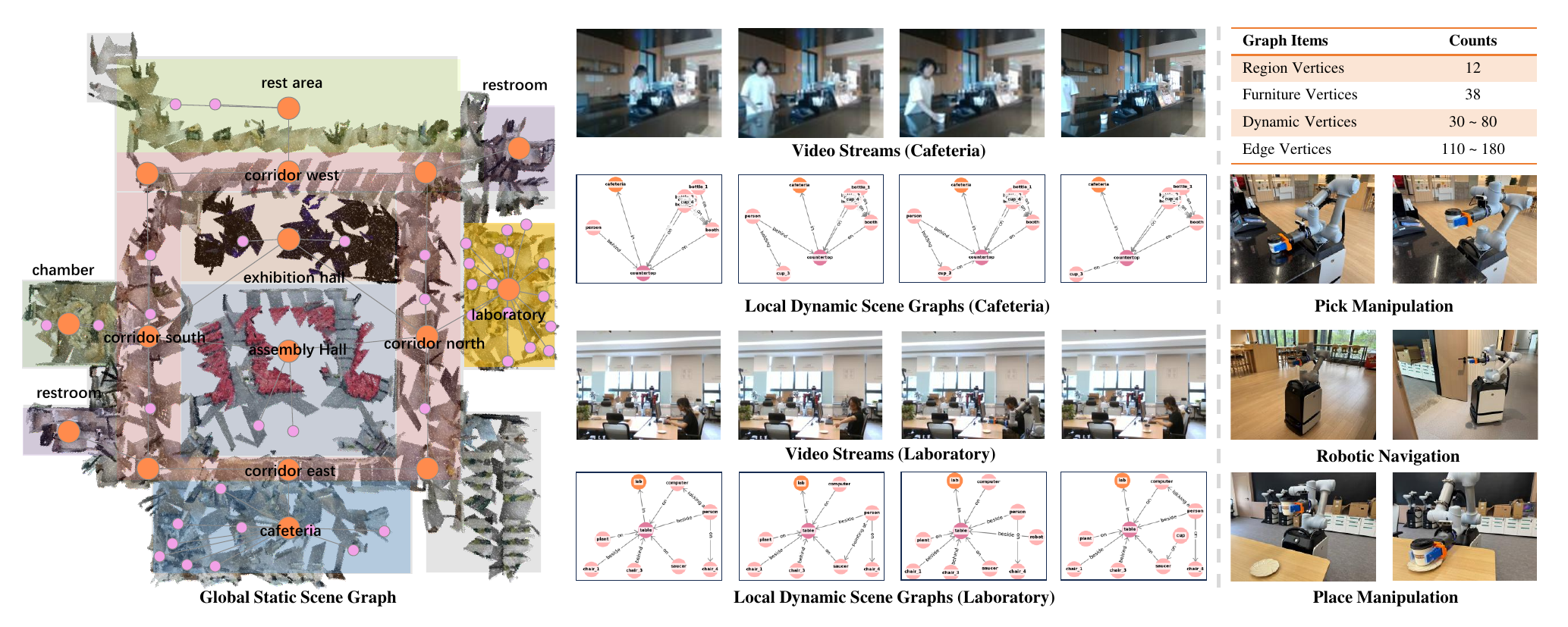}
\vspace{-0.25 in}
\caption{\textbf{Example of the generated hierarchical dynamic scene graphs.} We show the detailed evolving process of dynamic subgraphs in the cafeteria and laboratory. On the right, we show the quantity counts of the graph vertices and edges during the process. We further show the mobile manipulation demonstrations, including navigation, pick, and place tasks.}
\vspace{-0.15 in}
\label{fig::exp_total}
\end{figure}

\subsection{Graph Structure Evaluation}
Fig. \ref{fig::exp_total} shows a built example of the hierarchical dynamic scene graph in a campus building scenario, whose structure includes multiple rooms. The table shows the quantitative information of the graph vertices and edges. The evolving vertices and edges in the visualization reflect the system’s capacity to adaptively update scene representations in response to environmental dynamics, demonstrating our framework’s ability to maintain spatiotemporal coherence in dynamic settings.

\subsection{Robotic Deployment}

\begin{figure}[t]
\centering
\includegraphics[width=\linewidth]{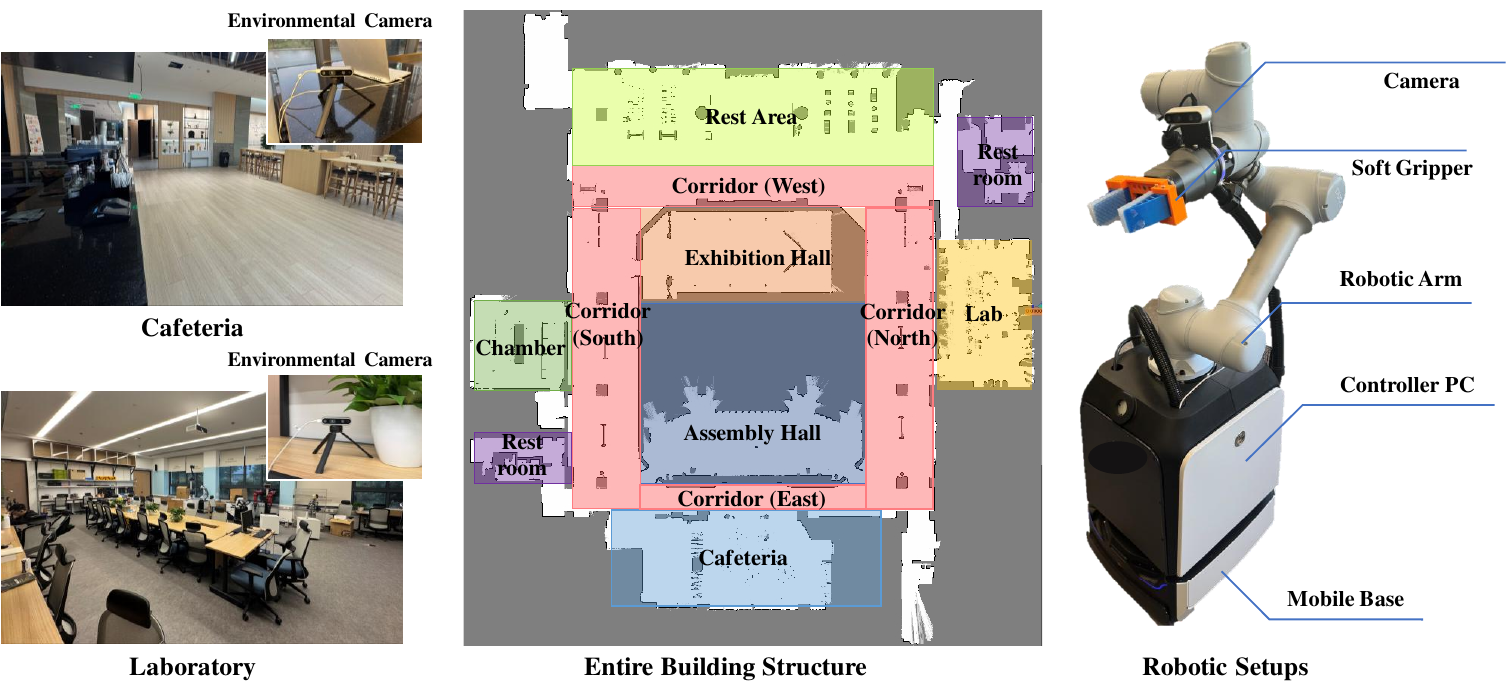}
\vspace{-0.25 in}
\caption{\textbf{Environmental and platform setups.} Environmental cameras are installed in the activity-critical regions. We show the top-down view of regions and a brief platform structure.}
\vspace{-0.2 in}
\label{fig::env_setup}
\end{figure}

\textbf{Setups.} 
For environmental setup, we install environmental cameras in activity-critical zones as shown in Fig. \ref{fig::env_setup}. All environmental cameras underwent ethical approval, with OpenCV-based tools automatically blurring faces for privacy compliance. For dynamic scene graph, we activate environmental cameras to record at 5 Hz. The environmental camera pose is aligned to the unified coordinates by feeding its records to GLOMAP~\cite{pan2024glomap} together with frames used to construct static graph. A $10\ \text{s}$ sliding window is applied to extract the most recent video sequences. Our robotic platform (see Fig. \ref{fig::env_setup}) integrates a SLAMTEC mobile base with a SIASUN manipulator. The base employs occupancy-based mapping and navigation, while the manipulator arm hosts a RealSense D435i camera. Full system calibration aligns the base, arm, and camera to unified coordinates. Detailed hardware specifications and environmental configurations are provided in the appendix.

\textbf{Demonstrations.}
We provide real-world robotic deployment to show the effectiveness of the proposed method. As shown in Fig. \ref{fig::exp_total}, equipped with navigation and pick-place skills, the robotic acts as a cafeteria assistant in our experiments. Taking coffee delivery as an example, when someone in the laboratory places a coffee order, the environmental camera in the cafeteria monitors the beverage preparation process. Once the brewed coffee is detected on the service counter, the robot is autonomously tasked with navigating from any location to retrieve the coffee and deliver it to the laboratory. The entire workflow is performed autonomously, without manual assistance.

\subsection{Effectiveness of Dynamic Components}
\textbf{Relation Prediction.}
For local dynamic scene graph generation, we conduct experiments on established public datasets, including Ego4D~\cite{grauman2022ego4d}, Epic-Kitchens~\cite{damen2018epickitchen}, and VIDOR~\cite{shang2019vidor} benchmarks originally utilized in PVSG~\cite{yang2023pvsg}. Unlike PVSG’s ResNet-50 backbone, our implementation adopts the ConvNeXt-Large CLIP backbone pre-trained on LAION-2B, following the FCCLIP~\cite{yu2023fcclip} framework. We directly leverage FC-CLIP’s inference strategy without fine-tuning to obtain segmentation results. For segment features tracking and relation prediction, we retain PVSG’s methodology: UniTrack~\cite{wang2021unitrack} associates cross-frame segmentation instances, while relation predictions derive from the candidate pair filtering and classification pipeline detailed in Section \ref{method::local_graph} with a transformer encoder.

\begin{table}[tp]
\footnotesize
\centering
\setlength{\abovecaptionskip}{0.1cm}
\renewcommand{\arraystretch}{1.1}
\setlength{\tabcolsep}{3.2pt}
\begin{tabular}{lccccccc}
\toprule % Top horizontal line
\multirow{2}{*}{\textbf{Methods}} & \multicolumn{3}{c}{\textbf{In-vocabulary}} & & \multicolumn{3}{c}{\textbf{Open-vocabulary}} \\ \cline{2-4} \cline{6-8}
                      & R/mR@20 & R/mR@50 & R/mR@100 & & R/mR@20 & R/mR@50 & R/mR@100 \\ \midrule
3DSGG~\cite{wald2020learning}       & 3.37/1.73 & 3.56/1.89 & 4.52/2.27 & & 3.42/1.81 & 3.98/2.26 & 4.97/2.91 \\
PSG4D~\cite{yang2023psg4d}          & 6.15/3.46 & 6.58/4.04 & 6.83/4.51 & & 6.61/3.72 & 7.02/4.48 & 7.11/4.95 \\ \midrule
Ours(w/o CNN-CLIP)                  & 8.21/5.33 & 8.69/6.01 & 9.04/6.78 & & 8.60/5.63 & 8.89/5.66 & 9.14/6.26 \\
Ours(w/o relation pair prior)       & 6.12/3.41 & 7.73/5.31 & 8.02/6.54 & & 9.64/6.76 & 9.82/6.95 & 9.93/7.11 \\
\textbf{Hi-Dyna Graph(Ours)}             & \textbf{8.40/6.25} & \textbf{9.75/7.59} & \textbf{10.56/8.90} & & \textbf{11.52/8.68} & \textbf{11.91/8.84} & \textbf{12.24/9.07} \\
\bottomrule % Bottom horizontal line
\end{tabular}
\caption{\textbf{Quantitative comparison of relation prediction} results on OpenPVSG~\cite{yang2023pvsg} dataset. We separately compare the in-vocabulary results and open-vocabulary results. (w/o CNN-CLIP) indicates utilizing ViT-based backbone instead of CNN-based CLIP while (w/o relation pair prior) means the relatively static scene assumption is not employed.}
\vspace{-0.2 in}
\label{tab::exp_2}
\end{table}

We evaluate the relation prediction results and separately compare the in-vocabulary results and open-vocabulary results. The in-vocabulary results consider a prediction as accurate only if the subject category, object category, and relation is exactly same as the GT label while the open-vocabulary results consider a prediction as accurate if it predicts a reasonable result similar to the GT (e.g. person sitting on sofa v.s. adult on couch). As shown is Tab. \ref{tab::exp_2}, our approach performs better than previous work, especially when evaluating the open-vocabulary results.

\begin{table}[tp]
\footnotesize
\centering
\setlength{\abovecaptionskip}{0.1cm}
\renewcommand{\arraystretch}{1.1}
\setlength{\tabcolsep}{1.6pt}
\begin{tabular}{ccccccccccccc}
\toprule % Top horizontal line
\multirow{2}{*}{\textbf{Graph Approach}} & \multirow{2}{*}{\textbf{Methods}} & \multicolumn{2}{c}{\textbf{0 min}} & & \multicolumn{2}{c}{\textbf{10 min}} & & \multicolumn{2}{c}{\textbf{20 min}} & & \multicolumn{2}{c}{\textbf{30 min}} \\ \cline{3-4} \cline{6-7} \cline{9-10} \cline{12-13}
                    & & V. Acc. & E. Acc. & & V. Acc. & E. Acc. & & V. Acc. & E. Acc. & & V. Acc. & E. Acc.  \\ \hline
\multirow{3}{*}{\makecell{Static \\ Built-from-Scratch}} & ConceptGraph*~\cite{gu2024conceptgraphs} & 0.68 & 0.90 & & 0.67 & 0.92 & & 0.65 & 0.89 & & 0.69 & 0.87  \\
                                         & HOV-SG~\cite{werby23hovsg}                               & 0.74 & \textbf{0.94} & & \textbf{0.72} & \textbf{\underline{0.96}} & & 0.70 & \textbf{0.95} & & \textbf{\underline{0.74}} & \textbf{0.95}  \\
                                         & Topo-Field~\cite{hou2025topofield}                       & \textbf{\underline{0.77}} & \textbf{\underline{0.96}} & & 0.69 & \textbf{0.93} & & \textbf{\underline{0.75}} & \textbf{\underline{0.96}} & & \textbf{0.72} & \textbf{\underline{0.96}}  \\ \midrule
Dynamic Updating &                         \textbf{Hi-Dyna (Ours)}                                  & \textbf{0.76} & 0.92 & & \textbf{\underline{0.73}} & 0.90 & & \textbf{0.71} & 0.94 & & \textbf{0.72} & \textbf{0.95}  \\
% \midrule % In-table horizontal line
% \midrule % In-table horizontal line
% Text queries               & \multicolumn{2}{c}{100} & \multicolumn{2}{c}{100} & \multicolumn{2}{c}{60} & \multicolumn{2}{c}{60} \\
\bottomrule % Bottom horizontal line
\end{tabular}
\caption{\textbf{Quantitative comparison of multilevel dynamic graph structure} at each time interval step against static methods as time passes by (evaluated right at once, after 10 min, after 20min, and after 30 min). The V. Acc. stands for the vertices accuracy and E. Acc. stands for edges accuracy. ConceptGraph*~\cite{gu2024conceptgraphs} indicates that, because ConceptGraph lacks explicit scene layout modeling, we substitute its layout topology with GT annotations for fair comparison. Bold and underlined numbers indicate the highest accuracy, while bold numbers represent the second highest.}
\label{tab::exp_3}
\vspace{-0.25in}
\end{table}

\textbf{Evaluation on Dynamic Components.}
To further validate the efficacy of our scene graph construction framework, we conduct comparative evaluations against existing static scene graph methods with the following strategy. Our multilevel dynamic scene graph is evaluated every 1 min, the graph is updated with a 10 s sliding window. while baseline methods reconstruct static scene graphs from scratch with full image sequences collected at every 1 min interval. Evaluation metrics focus on the accuracy of vertices and edges. Results in Tab. \ref{tab::exp_3} indicate that our dynamic update method shows competitive performance on the updated graph structure even compared against static graph generation methods which build the graph from scratch at each time interval step.

\subsection{Ablations}
\textbf{Static Objects Filter.} We ablate the relatively static objects filter to show its efficacy in helping construct robust static components. We employ HOV-SG~\cite{werby23hovsg} and Topo-Field~\cite{hou2025topofield} as baselines. The employed HOV-SG~\cite{werby23hovsg} considers ViT-H-14 as a CLIP~\cite{clip} backbone as described in its paper, while Topo-Field~\cite{hou2025topofield} leverages a ViT-B-32 CLIP~\cite{clip} backbone and an S-BERT~\cite{sentence-bert} encoder all-mpnet-base-v2. Based on these two approaches, the employed backbone in our method keeps align with them separately. The volumn threshold $\text{v}_{thr}$ described in Section \ref{method::global_graph} is set to $2\ \text{m}^3$. 
The dataset was collected from a single floor of a campus academic building, covering diverse functional zones. We recorded multiple sequences across varying times of day to capture environmental variations with various activities.The evaluated metric is the vertices precision that indicates the accuracy of the established vertice as described in ConceptGraph~\cite{gu2024conceptgraphs}. A vertice is considered as correct if the predicted label is correct and the overlap of predicted bounding-box and GT one is more than 60\%.

\begin{figure}[t]
\centering
\includegraphics[width=\linewidth]{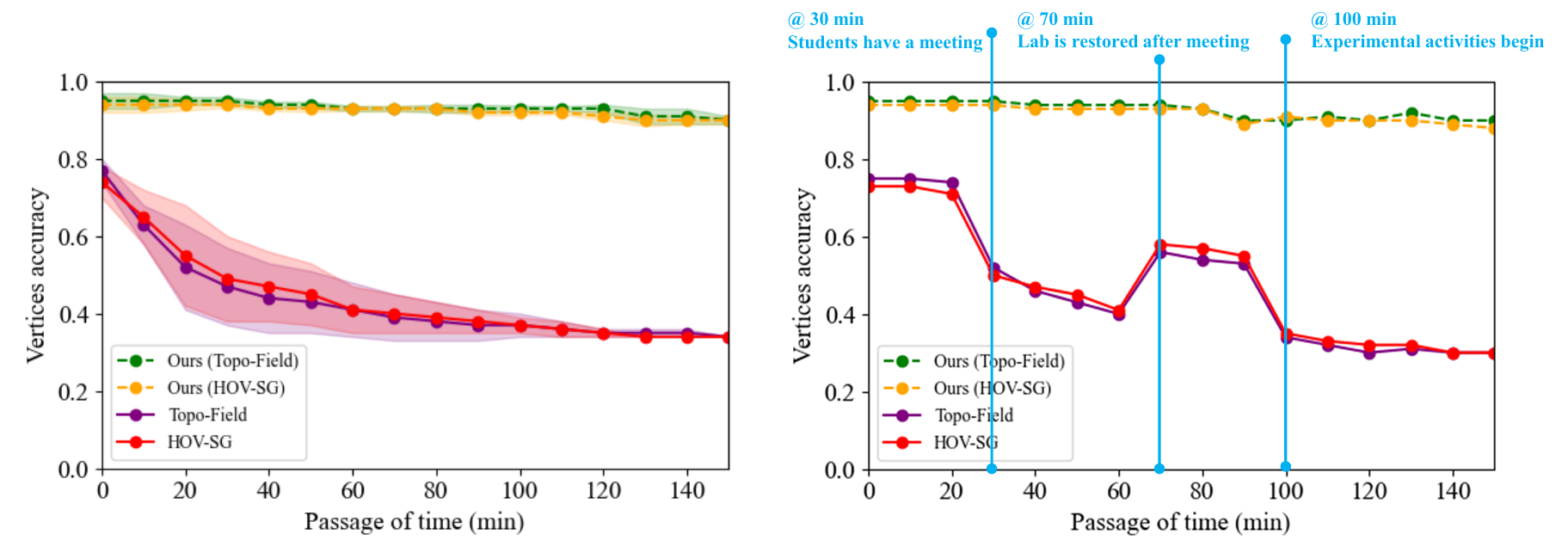}
\vspace{-0.2 in}
\caption{\textbf{Comparison of static graph vertices accuracy as time goes by.} The left plot illustrates the temporal variations in vertices accuracy across static scene graphs constructed using different methods from multiple video sequences. The right plot demonstrates the time-dependent accuracy of scene graphs in a laboratory over an extended period, with key timestamps annotated to highlight human activities that significantly impact the constructed vertices.}
\vspace{-0.2 in}
\label{fig::exp1}
\end{figure}

As time goes by, we continually compare the accuracy of the constructed vertices. The results shown in Fig. \ref{fig::exp1} indicate that our method effectively identifies and prioritizes stable scene components, specifically, the spatial distribution of functional regions and fixed macro-objects (e.g., furniture). Over extended temporal intervals, Hi-Dyna graph exhibits minimal degradation from environmental dynamics, validating its robustness to transient perturbations in human-centric settings. The results also demonstrate that human activities have a significant impact on graphs built by existing methods.

\textbf{Dynamic Graph Generation} As shown in Tab. \ref{tab::exp_2}, we ablate on the segmentation backbone and the pair proposal strategy during the dynamic graph generation process. Ours(w/o CNN-CLIP) means we ablate the introduced CNN-based CLIP encoder and employ a ViT-based one. Ours(w/o relation pair prior) means we do not introduce the relatively static scene assumption that considers human as more likely to be the subject and large furniture as more likely to be the object. The results show that these strategies effectively improve performance.

\section{Conclusion and Limitations}
This work introduces Hi-Dyna Graph for embodied autonomy in human-centric environments. By conducting complex experiments and real-world deployment we demonstrate promising effectiveness, however, several limitations warrant discussion. 1) While our graph architecture balances efficiency and dynamism, the sliding window for dynamic updates may introduce latency in rapidly evolving scenarios. Our current realization processes at a fixed frequency but has a potential to reach the real-time application by improving the segmentation, tracking, and relation prediction models. 2) The real-world robotic deployment is realized with a junior approach by employing simple polices and given skills to ground the scene knowledge and representation. However, there exist more powerful vision-language approaches which can enable the robotic with more robust and efficient abilities.

\bibliography{example}

\begin{thebibliography}{10}

\bibitem{werby23hovsg}
Abdelrhman Werby, Chenguang Huang, Martin Büchner, Abhinav Valada, and Wolfram Burgard.
\newblock Hierarchical open-vocabulary 3d scene graphs for language-grounded robot navigation.
\newblock {\em Robotics: Science and Systems}, 2024.

\bibitem{Maggio2024Clio}
Dominic Maggio, Yun Chang, Nathan Hughes, Matthew Trang, Dan Griffith, Carlyn Dougherty, Eric Cristofalo, Lukas Schmid, and Luca Carlone.
\newblock Clio: Real-time task-driven open-set 3d scene graphs.
\newblock {\em IEEE Robotics and Automation Letters}, 9(10):8921--8928, 2024.

\bibitem{hou2025topofield}
Jiawei Hou, Wenhao Guan, Longfei Liang, Jianfeng Feng, Xiangyang Xue, and Taiping Zeng.
\newblock Topo-field: Topometric mapping with brain-inspired hierarchical layout-object-position fields.
\newblock {\em IEEE Robotics and Automation Letters}, 10(6):5385--5392, 2025.

\bibitem{blochliger2018topomap}
Fabian Blochliger, Marius Fehr, Marcin Dymczyk, Thomas Schneider, and Rol Siegwart.
\newblock Topomap: Topological mapping and navigation based on visual slam maps.
\newblock In {\em 2018 IEEE International Conference on Robotics and Automation (ICRA)}, pages 3818--3825. IEEE, 2018.

\bibitem{gomez2020hybrid}
Clara Gomez, Marius Fehr, Alex Millane, Alejandra~C Hernandez, Juan Nieto, Ramon Barber, and Roland Siegwart.
\newblock Hybrid topological and 3d dense mapping through autonomous exploration for large indoor environments.
\newblock In {\em 2020 IEEE International Conference on Robotics and Automation (ICRA)}, pages 9673--9679. IEEE, 2020.

\bibitem{topo1}
Qiwen Zhang.
\newblock {\em Autonomous indoor exploration and mapping using hybrid metric/topological maps}.
\newblock McGill University (Canada), 2015.

\bibitem{topo2}
Qiwen Zhang, Ioannis Rekleitis, and Gregory Dudek.
\newblock Uncertainty reduction via heuristic search planning on hybrid metric/topological map.
\newblock In {\em 2015 12th Conference on Computer and Robot Vision}, pages 222--229. IEEE, 2015.

\bibitem{topo3}
Lu{\'\i}s Garrote, Cristiano Premebida, David Silva, and Urbano~J Nunes.
\newblock Hmaps-hybrid height-voxel maps for environment representation.
\newblock In {\em 2018 IEEE/RSJ International Conference on Intelligent Robots and Systems (IROS)}, pages 1197--1203. IEEE, 2018.

\bibitem{nerf}
Ben Mildenhall, Pratul~P. Srinivasan, Matthew Tancik, Jonathan~T. Barron, Ravi Ramamoorthi, and Ren Ng.
\newblock {NeRF}: Representing scenes as neural radiance fields for view synthesis.
\newblock In {\em The European Conference on Computer Vision (ECCV)}, 2020.

\bibitem{kerbl3Dgaussians}
Bernhard Kerbl, Georgios Kopanas, Thomas Leimk{\"u}hler, and George Drettakis.
\newblock 3d gaussian splatting for real-time radiance field rendering.
\newblock {\em ACM Transactions on Graphics}, 42(4), July 2023.

\bibitem{Wu20244DGS}
Guanjun Wu, Taoran Yi, Jiemin Fang, Lingxi Xie, Xiaopeng Zhang, Wei Wei, Wenyu Liu, Qi~Tian, and Xinggang Wang.
\newblock 4d gaussian splatting for real-time dynamic scene rendering.
\newblock In {\em Proceedings of the IEEE/CVF Conference on Computer Vision and Pattern Recognition (CVPR)}, pages 20310--20320, June 2024.

\bibitem{attal2023hyperreel}
Benjamin Attal, Jia-Bin Huang, Christian Richardt, Michael Zollhoefer, Johannes Kopf, Matthew O'Toole, and Changil Kim.
\newblock {HyperReel}: {H}igh-fidelity {6-DoF} video with ray-conditioned sampling.
\newblock In {\em CVPR}, 2023.

\bibitem{hexplane}
Ang Cao and Justin Johnson.
\newblock Hexplane: A fast representation for dynamic scenes.
\newblock In {\em 2023 IEEE/CVF Conference on Computer Vision and Pattern Recognition (CVPR)}, pages 130--141, 2023.

\bibitem{kplane}
Sara Fridovich-Keil, Giacomo Meanti, Frederik~Rahbæk Warburg, Benjamin Recht, and Angjoo Kanazawa.
\newblock K-planes: Explicit radiance fields in space, time, and appearance.
\newblock In {\em 2023 IEEE/CVF Conference on Computer Vision and Pattern Recognition (CVPR)}, pages 12479--12488, 2023.

\bibitem{yang2023pvsg}
Jingkang Yang, Wenxuan Peng, Xiangtai Li, Zujin Guo, Liangyu Chen, Bo~Li, Zheng Ma, Kaiyang Zhou, Wayne Zhang, Chen~Change Loy, and Ziwei Liu.
\newblock Panoptic video scene graph generation.
\newblock In {\em CVPR}, 2023.

\bibitem{yang2023psg4d}
Jingkang Yang, Jun Cen, Wenxuan Peng, Fangzhou Liu, Shuai amd~Hong, Xiangtai Li, Kaiyang Zhou, Qifeng Chen, and Ziwei Liu.
\newblock 4d panoptic scene graph generation.
\newblock In {\em NeurIPS}, 2023.

\bibitem{tnerf}
Chen Gao, Ayush Saraf, Johannes Kopf, and Jia-Bin Huang.
\newblock Dynamic view synthesis from dynamic monocular video.
\newblock In {\em 2021 IEEE/CVF International Conference on Computer Vision (ICCV)}, pages 5692--5701, 2021.

\bibitem{li2021neural}
Zhengqi Li, Simon Niklaus, Noah Snavely, and Oliver Wang.
\newblock Neural scene flow fields for space-time view synthesis of dynamic scenes.
\newblock In {\em Proceedings of the IEEE/CVF Conference on Computer Vision and Pattern Recognition}, pages 6498--6508, 2021.

\bibitem{li2022neural}
Tianye Li, Mira Slavcheva, Michael Zollhoefer, Simon Green, Christoph Lassner, Changil Kim, Tanner Schmidt, Steven Lovegrove, Michael Goesele, Richard Newcombe, et~al.
\newblock Neural 3d video synthesis from multi-view video.
\newblock In {\em Proceedings of the IEEE/CVF conference on computer vision and pattern recognition}, pages 5521--5531, 2022.

\bibitem{du2021neural}
Yilun Du, Yinan Zhang, Hong-Xing Yu, Joshua~B Tenenbaum, and Jiajun Wu.
\newblock Neural radiance flow for 4d view synthesis and video processing.
\newblock In {\em 2021 IEEE/CVF International Conference on Computer Vision (ICCV)}, pages 14304--14314. IEEE Computer Society, 2021.

\bibitem{park2021hypernerf}
Keunhong Park, Utkarsh Sinha, Peter Hedman, Jonathan~T Barron, Sofien Bouaziz, Dan~B Goldman, Ricardo Martin-Brualla, and Steven~M Seitz.
\newblock Hypernerf: a higher-dimensional representation for topologically varying neural radiance fields.
\newblock {\em ACM Transactions on Graphics (TOG)}, 40(6):1--12, 2021.

\bibitem{park2021nerfies}
Keunhong Park, Utkarsh Sinha, Jonathan~T Barron, Sofien Bouaziz, Dan~B Goldman, Steven~M Seitz, and Ricardo Martin-Brualla.
\newblock Nerfies: Deformable neural radiance fields.
\newblock In {\em Proceedings of the IEEE/CVF international conference on computer vision}, pages 5865--5874, 2021.

\bibitem{yang2023deformable3dgs}
Ziyi Yang, Xinyu Gao, Wen Zhou, Shaohui Jiao, Yuqing Zhang, and Xiaogang Jin.
\newblock Deformable 3d gaussians for high-fidelity monocular dynamic scene reconstruction.
\newblock {\em arXiv preprint arXiv:2309.13101}, 2023.

\bibitem{spacegs}
Zhan Li, Zhang Chen, Zhong Li, and Yi~Xu.
\newblock Spacetime gaussian feature splatting for real-time dynamic view synthesis.
\newblock In {\em Proceedings of the IEEE/CVF Conference on Computer Vision and Pattern Recognition (CVPR)}, pages 8508--8520, June 2024.

\bibitem{yu2023fcclip}
Qihang Yu, Ju~He, Xueqing Deng, Xiaohui Shen, and Liang-Chieh Chen.
\newblock Convolutions die hard: Open-vocabulary segmentation with single frozen convolutional clip.
\newblock In {\em NeurIPS}, 2023.

\bibitem{wang2021unitrack}
Z~Wang, H~Zhao, Y~Li, S~Wang, P~Torr, and L~Bertinetto.
\newblock Do different tracking tasks require different appearance models?
\newblock In {\em 2021 Conference on Neural Information Processing Systems (NeurIPS)}, 2021.

\bibitem{wang2024pair}
Jinghao Wang, Zhengyu Wen, Xiangtai Li, Zujin Guo, Jingkang Yang, and Ziwei Liu.
\newblock Pair then relation: Pair-net for panoptic scene graph generation.
\newblock {\em IEEE Transactions on Pattern Analysis and Machine Intelligence}, 2024.

\bibitem{hou2025elazson}
Jiawei Hou, Yuting Xiao, Xiangyang Xue, and Taiping Zeng.
\newblock Ela-zson: Efficient layout-aware zero-shot object navigation agent with hierarchical planning.
\newblock {\em arXiv preprint arXiv:2505.06131}, 2025.

\bibitem{wang2024polaris}
Tianyu Wang, Haitao Lin, Junqiu Yu, and Yanwei Fu.
\newblock Polaris: Open-ended interactive robotic manipulation via syn2real visual grounding and large language models.
\newblock In {\em 2024 IEEE/RSJ International Conference on Intelligent Robots and Systems (IROS)}, pages 9676--9683. IEEE, 2024.

\bibitem{pan2024glomap}
Linfei Pan, Daniel Barath, Marc Pollefeys, and Johannes~Lutz Sch\"{o}nberger.
\newblock {Global Structure-from-Motion Revisited}.
\newblock In {\em European Conference on Computer Vision (ECCV)}, 2024.

\bibitem{grauman2022ego4d}
Kristen Grauman, Andrew Westbury, Eugene Byrne, Zachary Chavis, Antonino Furnari, Rohit Girdhar, Jackson Hamburger, Hao Jiang, Miao Liu, Xingyu Liu, et~al.
\newblock Ego4d: Around the world in 3,000 hours of egocentric video.
\newblock In {\em Proceedings of the IEEE/CVF conference on computer vision and pattern recognition}, pages 18995--19012, 2022.

\bibitem{damen2018epickitchen}
Dima Damen, Hazel Doughty, Giovanni~Maria Farinella, Sanja Fidler, Antonino Furnari, Evangelos Kazakos, Davide Moltisanti, Jonathan Munro, Toby Perrett, Will Price, et~al.
\newblock Scaling egocentric vision: The epic-kitchens dataset.
\newblock In {\em Proceedings of the European conference on computer vision (ECCV)}, pages 720--736, 2018.

\bibitem{shang2019vidor}
Xindi Shang, Donglin Di, Junbin Xiao, Yu~Cao, Xun Yang, and Tat-Seng Chua.
\newblock Annotating objects and relations in user-generated videos.
\newblock In {\em Proceedings of the 2019 on International Conference on Multimedia Retrieval}, pages 279--287, 2019.

\bibitem{wald2020learning}
Johanna Wald, Helisa Dhamo, Nassir Navab, and Federico Tombari.
\newblock Learning 3d semantic scene graphs from 3d indoor reconstructions.
\newblock In {\em Proceedings of the IEEE/CVF Conference on Computer Vision and Pattern Recognition}, pages 3961--3970, 2020.

\bibitem{gu2024conceptgraphs}
Qiao Gu, Ali Kuwajerwala, Sacha Morin, Krishna~Murthy Jatavallabhula, Bipasha Sen, Aditya Agarwal, Corban Rivera, William Paul, Kirsty Ellis, Rama Chellappa, et~al.
\newblock Conceptgraphs: Open-vocabulary 3d scene graphs for perception and planning.
\newblock In {\em 2024 IEEE International Conference on Robotics and Automation (ICRA)}, pages 5021--5028. IEEE, 2024.

\bibitem{clip}
Alec Radford, Jong~Wook Kim, Chris Hallacy, Aditya Ramesh, Gabriel Goh, Sandhini Agarwal, Girish Sastry, Amanda Askell, Pamela Mishkin, Jack Clark, et~al.
\newblock Learning transferable visual models from natural language supervision.
\newblock In {\em International conference on machine learning}, pages 8748--8763. PMLR, 2021.

\bibitem{sentence-bert}
Nils Reimers and Iryna Gurevych.
\newblock Sentence-bert: Sentence embeddings using siamese bert-networks.
\newblock In {\em Proceedings of the 2019 Conference on Empirical Methods in Natural Language Processing and the 9th International Joint Conference on Natural Language Processing (EMNLP-IJCNLP)}, pages 3982--3992, Hong Kong, China, November 2019. Association for Computational Linguistics.

\bibitem{detic}
Xingyi Zhou, Rohit Girdhar, Armand Joulin, Philipp Kr{\"a}henb{\"u}hl, and Ishan Misra.
\newblock Detecting twenty-thousand classes using image-level supervision.
\newblock In {\em ECCV}, 2022.

\end{thebibliography}

%%%%%%%%%%%%%%%%%%%%%%%%%%%%%%%%%%%%%%%%%%%%%%%%%%%%%%%%%%%%
\newpage
\appendix

\section{Appendix / supplemental material}
\renewcommand\thefigure{\Alph{section}\arabic{figure}}
\setcounter{figure}{0}

\subsection{Environmental Setups}
\begin{figure}[h]
\centering
\includegraphics[width=\linewidth]{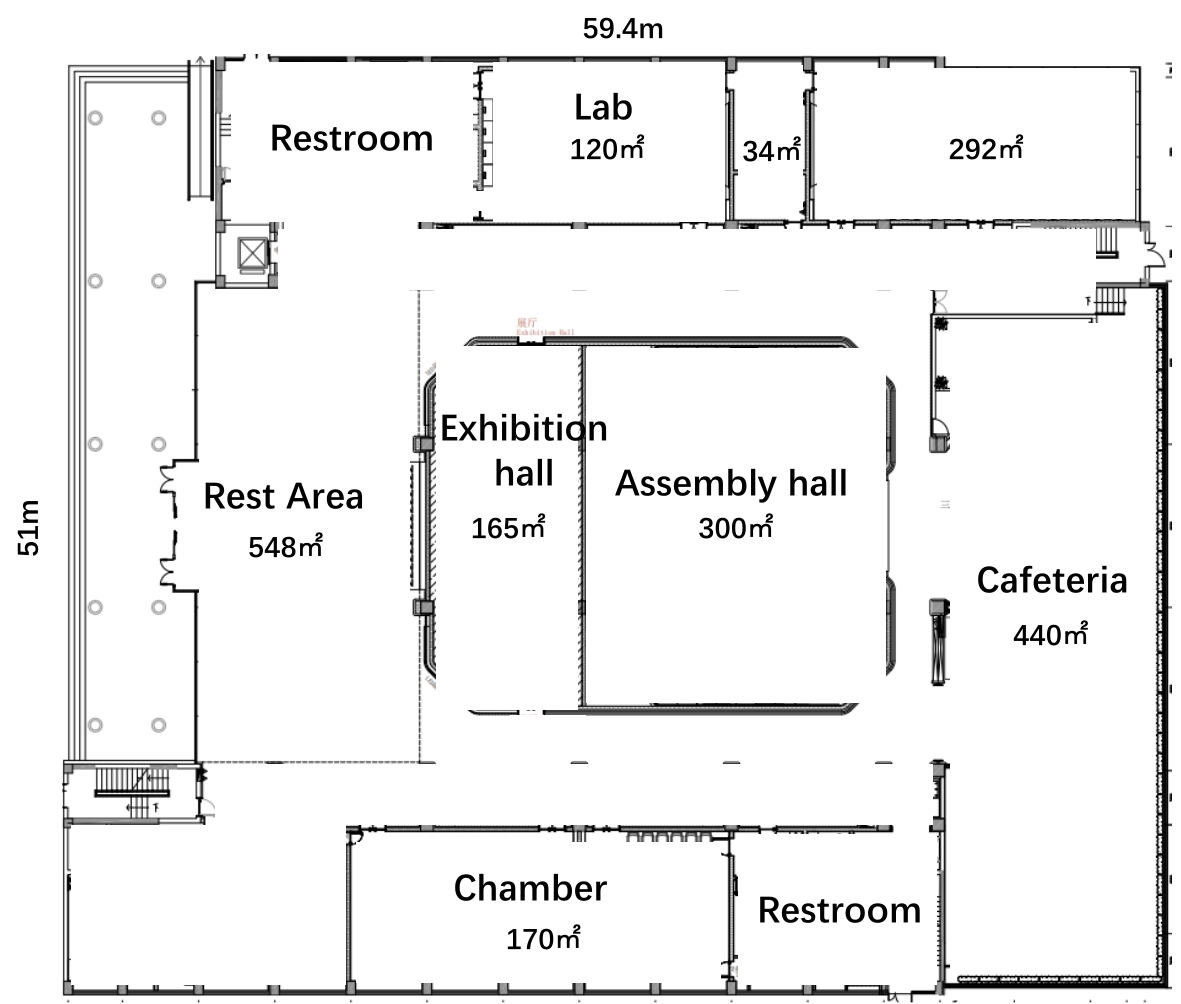}
% \vspace{-0.2 in}
\caption{The top-down view of the campus building structure with the exact scale.}
\label{fig::app::structure}
\end{figure}

The environment we deploy our robotics and conduct experiments is a multi-room indoor scenario of a campus building about $3029.4 m^2$, consisting of a laboratory area of about $120 m^2$, a cafeteria of about $440 m^2$, an exhibition hall of about $164.8 m^2$, an assembly hall of about $300 m^2$, a rest area of about $548 m^2$, a chamber of about $170 m^2$, corridors of about $121.2 m$ long, two restrooms, and several offices. Target manipulating objects in our experiments are chosen from the main function areas. The environmental camera is set at the cafeteria and the laboratory. The overview of the scene with exact scale is shown in Fig. \ref{fig::app::structure}.

\subsection{Robotic Setups}
\begin{figure}[h]
\centering
\includegraphics[width=\linewidth]{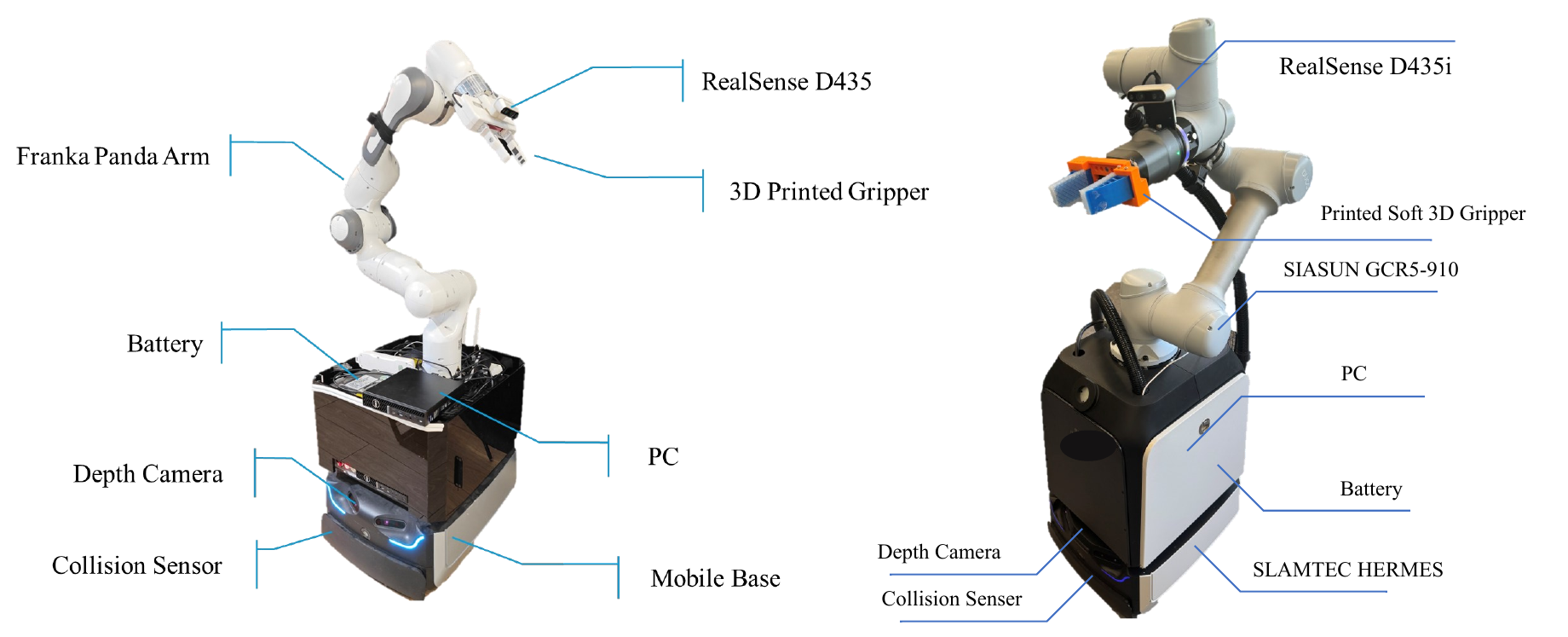}
% \vspace{-0.2 in}
\caption{The hardware platform of our employed robotic mobile manipulators.}
\label{fig::app::robot_setup}
\end{figure}

The construction of the robotic embodiment is shown in Fig.\ref{fig::app::robot_setup}. We employ two types of platforms. Both mainly consist of a mobile base, a robotic arm, a Realsense D435 camera attached to the robotic arm end effector, a battery unit, and a PC. The camera is calibrated with the robotic arm base with the easy-hand-eye package. The PC is used to take control of the mobile base and get the RGB-D frame from the camera. The transformation from the arm base to the mobile base coordinates center is considered to align the RGB-D frames to the base coordinates. The maximum velocity of the mobile base is set to 1$m/s$. For the mobile base, we employ the SLAMTEC Hermes, equipped with a laser radar for simplified localization and obstacle avoidance. For robotic arm, one platform utilizes the Franka panda arm, the other utilizes the SIASUN GCR5-910 arm. The graph construction, LLM reasoning, and object pose prediction algorithms are deployed on a PC equipped with and NVIDIA RTX 4090 GPU. The robotic mapping and localization, navigation, and manipulation processes are conducted on an embodied PC with an Intel i9-10885H CPU and GTX 1650ti GPU.

\textbf{Scene Graph Construction.}
We leverage a RealSense D435i RGB-D camera to capture frames of the whole scene and employ GLOMAP~\cite{pan2024glomap} to acquire the camera poses. The global static graph generation follows the description in Section \ref{method::global_graph}. For dynamic scene graph, we activate environmental cameras to record at 5 Hz. The environmental camera pose is aligned to the unified coordinates by feeding its records to GLOMAP~\cite{pan2024glomap} together with frames used to construct static graph. A $10\ \text{s}$ sliding window is applied to extract the most recent video sequences, from which dynamic relation graphs are generated and updated as in Section \ref{method::local_graph}.

\textbf{Robotic Manipulation.}
To enable the robot to execute embodied tasks, all feasible skills are formulated as combinations of navigation and manipulator-based pick-and-place actions. For navigation, we pre-map the environment using the LiDAR on the mobile base to construct a 2D occupancy map with 5 cm resolution, ensuring localization accuracy within $\pm5\ \text{cm}$. Waypoint-guided navigation is implemented via SLAMTEC’s proprietary API, adhering to their documented protocols. Upon reaching target locations, the robotic arm is maneuvered to position its end-effector-mounted camera at a downward-angled viewpoint ($\approx45^\circ $ tilt) for optimal object observation. Captured RGB-D images feed into a 6-DoF pose estimation pipeline, with the arm executing pose-guided grasping and placement.

\subsection{Model Setups}
At the static graph construction stage, 
\begin{itemize}
    \item For Topo-Field baseline, as described in their tutorial, CLIP with SwinB is employed in Detic~\cite{detic}, CLIP~\cite{clip} encoder is ViT-B/32 and Sentence-BERT~\cite{sentence-bert} encoder is all-mpnet-base-v2. The MHE has $18$ levels of grids and the dimension of each grid is $8$, with $log_2$ hash map size of $20$ and only 1 hidden MLP layer of size $600$. 
    \item For HOV-SG baseline, the CLIP backbone encoder is ViT-H-14 and the SAM encoder is $\text{ViT\_h\_4b}$. The voxel size is set as $0.02 m$. Other model parameters are kept align with the original setups.
\end{itemize}

\begin{figure}[h]
\centering
\includegraphics[width=\linewidth]{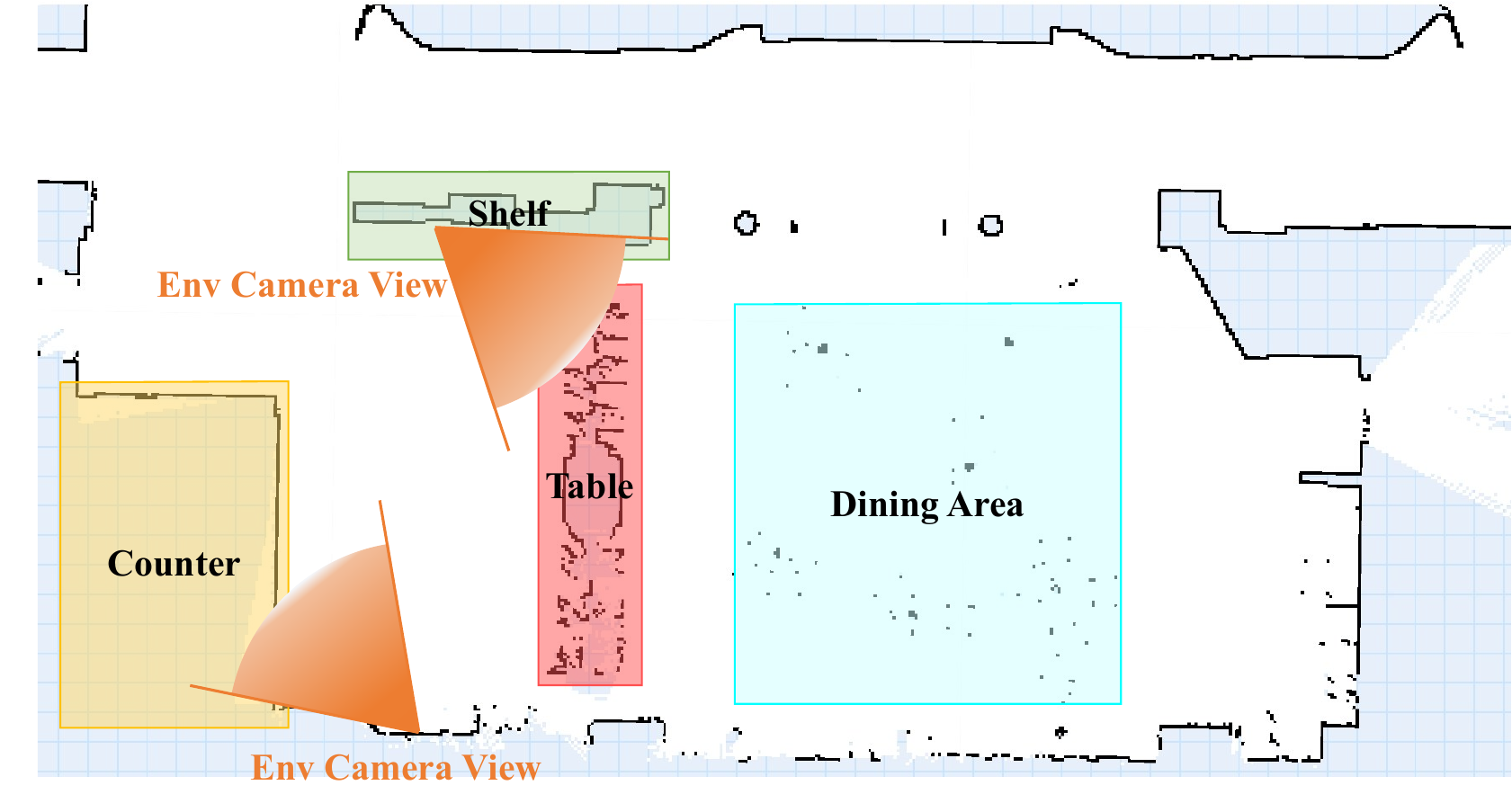}
% \vspace{-0.2 in}
\caption{The top-down view of layout and environmental cameras setups of the cafeteria area.}
\label{fig::app::cafe}
\end{figure}

At the dynamic graph generation stage, FC-CLIP is leveraged as segmentation baseline. The FC-CLIP employs ConvNeXt-L-d-320 as backbone which is pretrained on LAION-2B. The embedding dim is set as 768. Unitrack is employed to link segmentation results and features among frames to get the mask tubes and feature tubes. The track follows Unitrack's default settings (config/imagenet resnet18 s3 womotion.yaml) in their Github repository for Multi-Object Tracking and Segmentation (MOTS). The relation model training process follows the settings described in OpenPVSG, we employ the transformer relation head approach which performs optimal to others. As for the environmental cameras setup, we show a top-down view of the cafeteria as an example where we note the installation and view of the environmental cameras as shown in Fig. \ref{fig::app::cafe}.

\begin{figure}[h]
\centering
\includegraphics[width=\linewidth]{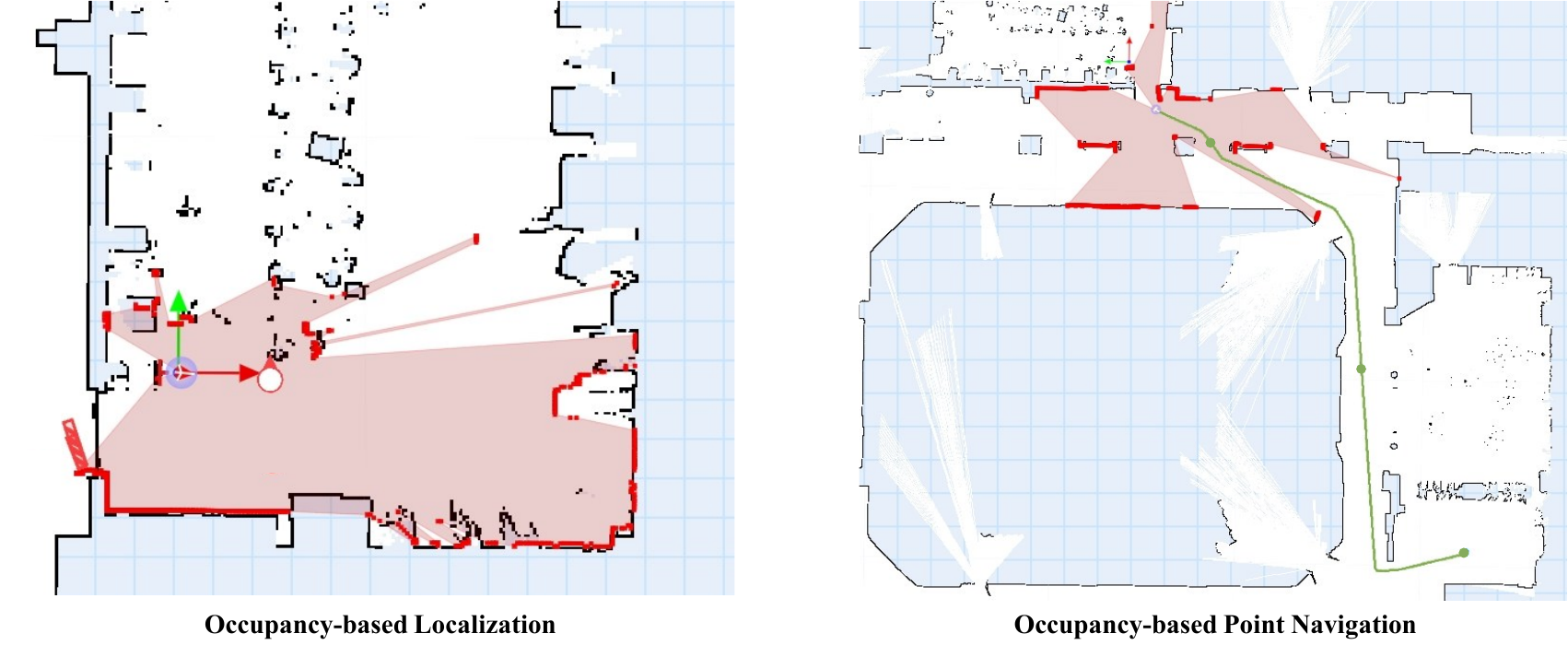}
% \vspace{-0.2 in}
\caption{Examples of the occupancy-based localization and navigation of our employed mobile base.}
\label{fig::app::occ}
\end{figure}

For navigation process, we build the 2D occupancy map based on the LiDAR input with the grid size of $5 cm$. Fig. \ref{fig::app::occ} shows an example of the built occupancy map and occupancy matching based localization and navigation. Consequently, the occupancy-based mapping and localization guides the pose-targeted navigation. We follow the API provided in the SLAMTEC tutorial.

\begin{figure}[h]
\centering
\includegraphics[width=\linewidth]{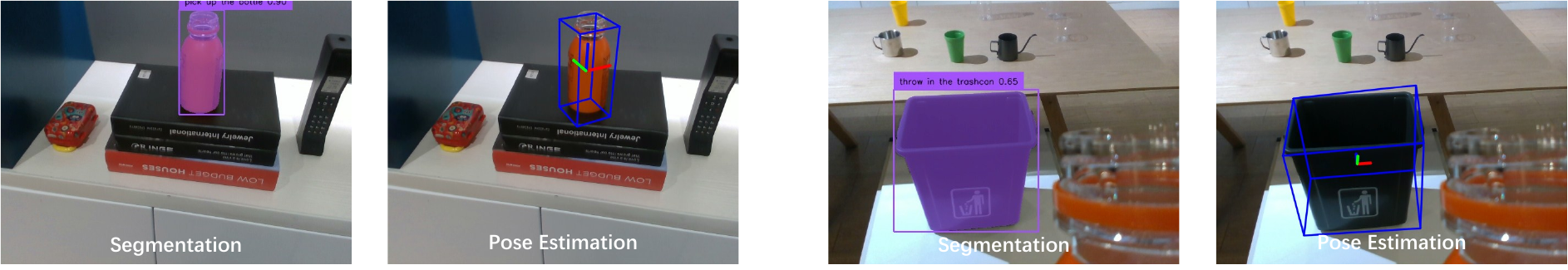}
% \vspace{-0.2 in}
\caption{Medium results of our employed object pose estimation approach. It takes RGB-D image and natural language text as input and outputs the localized object with its 6-DoF pose for manipulation.}
\label{fig::app::object_pose}
\end{figure}

For robotic arm manipulation, we employ Polaris as the pick-place task manipulator. It takes the observation and pick-place instruction as input, leverages Grounded Light HQSAM and MVPoseNet 6D to manage the instruction grounding, and drives the robotic arm to finish the 6-DoF pose guided tasks. Fig. \ref{fig::app::object_pose} shows the experimental medium results of the employed manipulation method, including the query and pose estimation result pairs.

\subsection{Task Videos}
We provide real-world robotic embodied deployment to show the effectiveness of the proposed method. The demonstration video shows the robotics have the ability to manage the autonomy in the campus building for given high-level tasks. In the shown video, the robotic is asked to help the coffee delivery task if someone in the lab orders a coffee from the cafeteria and help keep the laboratory and cafeteria clean. Two examples are given:

\begin{itemize}
    \item \textbf{Coffee Delivery.} As illustrated in the experiment section, The robotic is expected to help the autonomous coffee delivery process.
    \item \textbf{Tidy Up.} The environmental camera in the cafeteria detect the departure of patrons. The table is considered as in tidy-up needy if a person has been eating at the table and gone leaving things on the table which is not valuable. The top-down detailed map is shown in Fig. \ref{fig::app::cafe} of the cafeteria region.
\end{itemize}

\subsection{Failure Cases}
Our framework, while effective in many scenarios, exhibits two primary failure modes under specific conditions.

\textbf{Relation Prediction Failures.}
Relation prediction failures persist in complex or densely cluttered environments. Despite advances in scene graph generation, the relation prediction model occasionally produces inaccurate inferences or omits task-critical relationships—particularly when multiple humans interact with objects simultaneously. For instance, in crowded cafe settings, the system may fail to identify which patron has vacated their seat, leading to delayed or erroneous cleanup tasks. Similarly, if a prepared coffee is temporarily occluded (e.g., by a passing customer) when placed on the service counter, the robot may overlook the delivery trigger, leaving the item undelivered indefinitely.

\textbf{Error Recovery.}
Error recovery limitations arise during long-horizon task execution. The current pipeline lacks robust mechanisms to diagnose and recover from partial failures in subtask sequences. For example, if a coffee cup grasp attempt fails due to pose estimation inaccuracies, the system does not autonomously verify subtask completion or initiate recovery protocols (e.g., re-attempting the grasp or notifying a human operator). Instead, such errors propagate silently, often resulting in full task abandonment. This limitation stems from the absence of fine-grained state verification modules and closed-loop feedback during action execution.

\subsection{LLM Prompts}
This section provide a prompt example that is utilized to manage the robotic as a cafeteria assistant, following the approach illustrated in the methodology section as shown in \ref{prompt_listing}.

\begin{lstlisting}[caption=LLM reasoning prompts example, breaklines=true, breakatwhitespace=true, frame=single, basicstyle=\ttfamily\small,label=prompt_listing]
{
# ===== Task Context & Role Definition =====
TASK_CONTEXT = {
    "environment": "A university academic building floor containing 1) Cafe area 2) Classroom cluster 3) Faculty offices 4) Common spaces",
    "role": "Embodied service robot for campus cafe named 'CafeBot'",
    "primary_objective": "Handle delivery tasks between cafe counter and various destinations while maintaining spatial awareness",
    "operational_constraints": [
        "Must navigate through mixed pedestrian-robot traffic areas",
        "Service radius limited to same-floor locations",
        "Business hours: 8:00-18:00 weekdays",
        "Priority for hot beverage deliveries under 8-minute window"
    ]
}
# ===== Multi-Level Scene Understanding =====
SCENE_UNDERSTANDING = {
    "global_topological_map": {
        "structure": "Hierarchical graph with two layer abstraction",
        "layer1_nodes": {
            "regions": ["Cafe_Station", "Classroom_A1-A6", "Office_B1-B4", "Elevator_Lobby", "Storage_Closet"],
            "key_objects": ["Main_Counter", "Coffee_Machine", "Pickup_Desk", "Emergency_Exit"]
        },
        "layer1_edges": {
            "physical_connections": [
                ("Cafe_Station <-> Elevator_Lobby via North corridor (15m)"),
                ("Elevator_Lobby <-> Classroom_A1 via East hallway (20m)"),
                ("Cafe_Station <-> Office_B1 via West passage (12m)")
            ],
            "semantic_links": [
                ("Coffee_Machine located_in Cafe_Station"),
                ("Pickup_Desk adjacent_to Main_Counter")
            ]
        },
        "navigation_example": 
            """IF goal=Deliver to Classroom_A3:
            1. Query global topology
            2. Find Cafe_Station -> Elevator_Lobby -> Classroom_A1-A6 cluster
            3. Calculate shortest path avoiding crowded zones during class breaks
            4. Update path when detecting temporary obstruction"""
    },

    "local_relational_graph": {
        "dynamic_nodes": {
            "human": ["student", "professor", "staff", "visitor"],
            "objects": {
                "static": ["table", "door", "fire_extinguisher"],
                "dynamic": ["rolling_chair", "food_tray", "mobile_device"]
            }
        },
        "relationship_edges": {
            "spatial": ["near", "blocking", "approaching"],
            "functional": ["waiting_for", "handing_over", "using"],
            "temporal": ["recently_arrived", "about_to_leave"]
        },
        "reasoning_example":
            """WHEN DETECTED:
            - Barista human_node performing place_action(coffee_cup, pickup_counter)
            - Steam rising from cup_node
            THEN INFER:
            1. Coffee order ready for delivery
            2. Cup needs stabilization during transport
            3. Priority elevation if customer waiting_time > 5min"""
    }
}
# ===== Reasoning Chain ===== 
REASONING_MECHANISM = {
    "core_operation_chain": [
        "Navigate -> Grasp -> Navigate -> Place"
    ],
    
    "task_decomposition_examples": {
        "coffee_delivery": {
            "input": "Deliver coffee from counter to Classroom A3",
            "step_breakdown": [
                {"action": "Navigate", 
                 "params": {"target": "Main_Counter", "path_constraints": "avoid_peak_traffic"}},
                {"action": "Grasp", 
                 "params": {"object": "coffee_cup", "constraints": "tilt_angle<15_degrees"}},
                {"action": "Navigate", 
                 "params": {"target": "Classroom_A3", "risk_avoidance": "minimize_liquid_spillage"}},
                {"action": "Place",
                 "params": {"surface": "lectern_desk", "position_accuracy": "3cm"}}
            ]
        },
        
        "trash_retrieval": {
            "input": "Retrieve trash bin from Office B2",
            "step_breakdown": [
                {"action": "Navigate", 
                 "params": {"target": "Office_B2", "door_operation": "auto_door_activation"}},
                {"action": "Grasp",
                 "params": {"object": "trash_bin", "grip_mode": "cylindrical_grasp"}},
                {"action": "Navigate",
                 "params": {"target": "Storage_Closet", "payload_awareness": "center_of_gravity_compensation"}},
                {"action": "Place",
                 "params": {"surface": "recycling_zone", "orientation": "label_facing_outward"}}
            ]
        }
    },
}

# Update PROMPT_INSTRUCTIONS
PROMPT_INSTRUCTIONS += f"""

4. Basic Operation Reasoning Chain:
- Mandatory decomposition structure: {REASONING_MECHANISM['core_operation_chain']}
- Standard workflow example: {REASONING_MECHANISM['task_decomposition_examples']
['coffee_delivery']['step_breakdown']}
- Error recovery protocols: {REASONING_MECHANISM['error_recovery_protocols']
['grasp_failure'][0:2]}

Strictly prohibited:
- Introducing action types beyond grasp/place
- Adding object modification or complex interaction
- Executing non-navigation movement commands
"""
# New minimal example
"document_transfer": {
    "input": "Transfer documents from printer room to professor's office",
    "step_breakdown": [
        {"action": "Navigate", 
         "params": {"target": "Printer_Room", "elevator_usage": "priority_freight_elevator"}},
        {"action": "Grasp",
         "params": {"object": "document_stack", "pressure_control": "prevent_paper_crease"}},
        {"action": "Navigate",
         "params": {"target": "Professor_Office", "social_navigation": "avoid_private_conversation_areas"}},
        {"action": "Place",
         "params": {"surface": "incoming_tray", "alignment": "parallel_to_desk_edge"}}
    ]
}
# ===== Execution Protocol =====
PROMPT_INSTRUCTIONS = f"""
You are {TASK_CONTEXT['role']} operating in {TASK_CONTEXT['environment']}. Your core capabilities include:

1. Topological Navigation:
- Maintain mental map: {SCENE_UNDERSTANDING['global_topological_map']['structure']}
- Use regional connections like {SCENE_UNDERSTANDING['global_topological_map']
['layer1_edges']['physical_connections'][0]} for long-range planning
- Example: {SCENE_UNDERSTANDING['global_topological_map']
['navigation_example']}

2. Situational Reasoning:
- Track relationships: {SCENE_UNDERSTANDING['local_relational_graph']
['relationship_edges']['spatial']}
- Make inferences like {SCENE_UNDERSTANDING['local_relational_graph']
['reasoning_example']}
- Detect human activities (e.g., professor_chen approaching door -> hold door open)

3. Skill Orchestration:
- Compose primitives: {ROBOT_CAPABILITIES['primitive_skills'].keys()}
- Follow workflow: {ROBOT_CAPABILITIES['task_decomposition']['example_workflow']}

When receiving requests:
1. Parse request into semantic components
2. Cross-verify with spatial relationships
3. Generate executable skill sequence
4. Monitor environment changes for adaptation
"""
}
\end{lstlisting}

%%%%%%%%%%%%%%%%%%%%%%%%%%%%%%%%%%%%%%%%%%%%%%%%%%%%%%%%%%%%

\newpage
\section*{NeurIPS Paper Checklist}

\begin{enumerate}

\item {\bf Claims}
    \item[] Question: Do the main claims made in the abstract and introduction accurately reflect the paper's contributions and scope?
    \item[] Answer: \answerYes{} % Replace by \answerYes{}, \answerNo{}, or \answerNA{}.
    \item[] Justification: This paper claims that with the help of the proposed multilevel dynamic scene graph, robotics can reach better embodied autonomy. The claim is explained in detail in the abstract and introduction, and the main contributions are listed in the introduction
    \item[] Guidelines: 
    \begin{itemize}
        \item The answer NA means that the abstract and introduction do not include the claims made in the paper.
        \item The abstract and/or introduction should clearly state the claims made, including the contributions made in the paper and important assumptions and limitations. A No or NA answer to this question will not be perceived well by the reviewers. 
        \item The claims made should match theoretical and experimental results, and reflect how much the results can be expected to generalize to other settings. 
        \item It is fine to include aspirational goals as motivation as long as it is clear that these goals are not attained by the paper. 
    \end{itemize}

\item {\bf Limitations}
    \item[] Question: Does the paper discuss the limitations of the work performed by the authors?
    \item[] Answer: \answerYes{} % Replace by \answerYes{}, \answerNo{}, or \answerNA{}.
    \item[] Justification: The limitations of this work lie in the gap to the real-time deployment and the robotic embodied task realization approach. These are declared in the conclusion section.
    \item[] Guidelines:
    \begin{itemize}
        \item The answer NA means that the paper has no limitation while the answer No means that the paper has limitations, but those are not discussed in the paper. 
        \item The authors are encouraged to create a separate "Limitations" section in their paper.
        \item The paper should point out any strong assumptions and how robust the results are to violations of these assumptions (e.g., independence assumptions, noiseless settings, model well-specification, asymptotic approximations only holding locally). The authors should reflect on how these assumptions might be violated in practice and what the implications would be.
        \item The authors should reflect on the scope of the claims made, e.g., if the approach was only tested on a few datasets or with a few runs. In general, empirical results often depend on implicit assumptions, which should be articulated.
        \item The authors should reflect on the factors that influence the performance of the approach. For example, a facial recognition algorithm may perform poorly when image resolution is low or images are taken in low lighting. Or a speech-to-text system might not be used reliably to provide closed captions for online lectures because it fails to handle technical jargon.
        \item The authors should discuss the computational efficiency of the proposed algorithms and how they scale with dataset size.
        \item If applicable, the authors should discuss possible limitations of their approach to address problems of privacy and fairness.
        \item While the authors might fear that complete honesty about limitations might be used by reviewers as grounds for rejection, a worse outcome might be that reviewers discover limitations that aren't acknowledged in the paper. The authors should use their best judgment and recognize that individual actions in favor of transparency play an important role in developing norms that preserve the integrity of the community. Reviewers will be specifically instructed to not penalize honesty concerning limitations.
    \end{itemize}

\item {\bf Theory assumptions and proofs}
    \item[] Question: For each theoretical result, does the paper provide the full set of assumptions and a complete (and correct) proof?
    \item[] Answer: \answerNA{} % Replace by \answerYes{}, \answerNo{}, or \answerNA{}.
    \item[] Justification: This paper does not include theoretical results. We use experiments to prove that the proposed scene graph is effective and applicable.
    \item[] Guidelines:
    \begin{itemize}
        \item The answer NA means that the paper does not include theoretical results. 
        \item All the theorems, formulas, and proofs in the paper should be numbered and cross-referenced.
        \item All assumptions should be clearly stated or referenced in the statement of any theorems.
        \item The proofs can either appear in the main paper or the supplemental material, but if they appear in the supplemental material, the authors are encouraged to provide a short proof sketch to provide intuition. 
        \item Inversely, any informal proof provided in the core of the paper should be complemented by formal proofs provided in appendix or supplemental material.
        \item Theorems and Lemmas that the proof relies upon should be properly referenced. 
    \end{itemize}

    \item {\bf Experimental result reproducibility}
    \item[] Question: Does the paper fully disclose all the information needed to reproduce the main experimental results of the paper to the extent that it affects the main claims and/or conclusions of the paper (regardless of whether the code and data are provided or not)?
    \item[] Answer: \answerYes{} % Replace by \answerYes{}, \answerNo{}, or \answerNA{}.
    \item[] Justification: The detailed environmental setups, hardware setups, and model architecture are illustrated in detail and the methodology section includes the detail pipeline. The appendix also helps as completion.
    \item[] Guidelines:
    \begin{itemize}
        \item The answer NA means that the paper does not include experiments.
        \item If the paper includes experiments, a No answer to this question will not be perceived well by the reviewers: Making the paper reproducible is important, regardless of whether the code and data are provided or not.
        \item If the contribution is a dataset and/or model, the authors should describe the steps taken to make their results reproducible or verifiable. 
        \item Depending on the contribution, reproducibility can be accomplished in various ways. For example, if the contribution is a novel architecture, describing the architecture fully might suffice, or if the contribution is a specific model and empirical evaluation, it may be necessary to either make it possible for others to replicate the model with the same dataset, or provide access to the model. In general. releasing code and data is often one good way to accomplish this, but reproducibility can also be provided via detailed instructions for how to replicate the results, access to a hosted model (e.g., in the case of a large language model), releasing of a model checkpoint, or other means that are appropriate to the research performed.
        \item While NeurIPS does not require releasing code, the conference does require all submissions to provide some reasonable avenue for reproducibility, which may depend on the nature of the contribution. For example
        \begin{enumerate}
            \item If the contribution is primarily a new algorithm, the paper should make it clear how to reproduce that algorithm.
            \item If the contribution is primarily a new model architecture, the paper should describe the architecture clearly and fully.
            \item If the contribution is a new model (e.g., a large language model), then there should either be a way to access this model for reproducing the results or a way to reproduce the model (e.g., with an open-source dataset or instructions for how to construct the dataset).
            \item We recognize that reproducibility may be tricky in some cases, in which case authors are welcome to describe the particular way they provide for reproducibility. In the case of closed-source models, it may be that access to the model is limited in some way (e.g., to registered users), but it should be possible for other researchers to have some path to reproducing or verifying the results.
        \end{enumerate}
    \end{itemize}

\item {\bf Open access to data and code}
    \item[] Question: Does the paper provide open access to the data and code, with sufficient instructions to faithfully reproduce the main experimental results, as described in supplemental material?
    \item[] Answer: \answerYes{} % Replace by \answerYes{}, \answerNo{}, or \answerNA{}.
    \item[] Justification: We leverage open-access datasets and references to the open-access code base. Modifications are illustrated in the method section. For real-world deployment, the entire pipeline is described in detail.
    \item[] Guidelines:
    \begin{itemize}
        \item The answer NA means that paper does not include experiments requiring code.
        \item Please see the NeurIPS code and data submission guidelines (\url{https://nips.cc/public/guides/CodeSubmissionPolicy}) for more details.
        \item While we encourage the release of code and data, we understand that this might not be possible, so “No” is an acceptable answer. Papers cannot be rejected simply for not including code, unless this is central to the contribution (e.g., for a new open-source benchmark).
        \item The instructions should contain the exact command and environment needed to run to reproduce the results. See the NeurIPS code and data submission guidelines (\url{https://nips.cc/public/guides/CodeSubmissionPolicy}) for more details.
        \item The authors should provide instructions on data access and preparation, including how to access the raw data, preprocessed data, intermediate data, and generated data, etc.
        \item The authors should provide scripts to reproduce all experimental results for the new proposed method and baselines. If only a subset of experiments are reproducible, they should state which ones are omitted from the script and why.
        \item At submission time, to preserve anonymity, the authors should release anonymized versions (if applicable).
        \item Providing as much information as possible in supplemental material (appended to the paper) is recommended, but including URLs to data and code is permitted.
    \end{itemize}

\item {\bf Experimental setting/details}
    \item[] Question: Does the paper specify all the training and test details (e.g., data splits, hyperparameters, how they were chosen, type of optimizer, etc.) necessary to understand the results?
    \item[] Answer: \answerYes{} % Replace by \answerYes{}, \answerNo{}, or \answerNA{}.
    \item[] Justification: The implementation details, model architectures, setups are described in the main paper and also appendix sections.
    \item[] Guidelines:
    \begin{itemize}
        \item The answer NA means that the paper does not include experiments.
        \item The experimental setting should be presented in the core of the paper to a level of detail that is necessary to appreciate the results and make sense of them.
        \item The full details can be provided either with the code, in appendix, or as supplemental material.
    \end{itemize}

\item {\bf Experiment statistical significance}
    \item[] Question: Does the paper report error bars suitably and correctly defined or other appropriate information about the statistical significance of the experiments?
    \item[] Answer: \answerYes{} % Replace by \answerYes{}, \answerNo{}, or \answerNA{}.
    \item[] Justification: We provide error bars for the main result comparisons.
    \item[] Guidelines:
    \begin{itemize}
        \item The answer NA means that the paper does not include experiments.
        \item The authors should answer "Yes" if the results are accompanied by error bars, confidence intervals, or statistical significance tests, at least for the experiments that support the main claims of the paper.
        \item The factors of variability that the error bars are capturing should be clearly stated (for example, train/test split, initialization, random drawing of some parameter, or overall run with given experimental conditions).
        \item The method for calculating the error bars should be explained (closed form formula, call to a library function, bootstrap, etc.)
        \item The assumptions made should be given (e.g., Normally distributed errors).
        \item It should be clear whether the error bar is the standard deviation or the standard error of the mean.
        \item It is OK to report 1-sigma error bars, but one should state it. The authors should preferably report a 2-sigma error bar than state that they have a 96\% CI, if the hypothesis of Normality of errors is not verified.
        \item For asymmetric distributions, the authors should be careful not to show in tables or figures symmetric error bars that would yield results that are out of range (e.g. negative error rates).
        \item If error bars are reported in tables or plots, The authors should explain in the text how they were calculated and reference the corresponding figures or tables in the text.
    \end{itemize}

\item {\bf Experiments compute resources}
    \item[] Question: For each experiment, does the paper provide sufficient information on the computer resources (type of compute workers, memory, time of execution) needed to reproduce the experiments?
    \item[] Answer: \answerYes{} % Replace by \answerYes{}, \answerNo{}, or \answerNA{}.
    \item[] Justification: The computation setup is declared in the experiment section.
    \item[] Guidelines:
    \begin{itemize}
        \item The answer NA means that the paper does not include experiments.
        \item The paper should indicate the type of compute workers CPU or GPU, internal cluster, or cloud provider, including relevant memory and storage.
        \item The paper should provide the amount of compute required for each of the individual experimental runs as well as estimate the total compute. 
        \item The paper should disclose whether the full research project required more compute than the experiments reported in the paper (e.g., preliminary or failed experiments that didn't make it into the paper). 
    \end{itemize}
    
\item {\bf Code of ethics}
    \item[] Question: Does the research conducted in the paper conform, in every respect, with the NeurIPS Code of Ethics \url{https://neurips.cc/public/EthicsGuidelines}?
    \item[] Answer: \answerYes{} % Replace by \answerYes{}, \answerNo{}, or \answerNA{}.
    \item[] Justification: This research conform with the NeurIPS Code of Ethics.
    \item[] Guidelines:
    \begin{itemize}
        \item The answer NA means that the authors have not reviewed the NeurIPS Code of Ethics.
        \item If the authors answer No, they should explain the special circumstances that require a deviation from the Code of Ethics.
        \item The authors should make sure to preserve anonymity (e.g., if there is a special consideration due to laws or regulations in their jurisdiction).
    \end{itemize}

\item {\bf Broader impacts}
    \item[] Question: Does the paper discuss both potential positive societal impacts and negative societal impacts of the work performed?
    \item[] Answer: \answerNA{} % Replace by \answerYes{}, \answerNo{}, or \answerNA{}.
    \item[] Justification: This research focuses on improving the robotic autonomy, which is currently in the stage of pre-research of algorithms and has no social impact.
    \item[] Guidelines:
    \begin{itemize}
        \item The answer NA means that there is no societal impact of the work performed.
        \item If the authors answer NA or No, they should explain why their work has no societal impact or why the paper does not address societal impact.
        \item Examples of negative societal impacts include potential malicious or unintended uses (e.g., disinformation, generating fake profiles, surveillance), fairness considerations (e.g., deployment of technologies that could make decisions that unfairly impact specific groups), privacy considerations, and security considerations.
        \item The conference expects that many papers will be foundational research and not tied to particular applications, let alone deployments. However, if there is a direct path to any negative applications, the authors should point it out. For example, it is legitimate to point out that an improvement in the quality of generative models could be used to generate deepfakes for disinformation. On the other hand, it is not needed to point out that a generic algorithm for optimizing neural networks could enable people to train models that generate Deepfakes faster.
        \item The authors should consider possible harms that could arise when the technology is being used as intended and functioning correctly, harms that could arise when the technology is being used as intended but gives incorrect results, and harms following from (intentional or unintentional) misuse of the technology.
        \item If there are negative societal impacts, the authors could also discuss possible mitigation strategies (e.g., gated release of models, providing defenses in addition to attacks, mechanisms for monitoring misuse, mechanisms to monitor how a system learns from feedback over time, improving the efficiency and accessibility of ML).
    \end{itemize}
    
\item {\bf Safeguards}
    \item[] Question: Does the paper describe safeguards that have been put in place for responsible release of data or models that have a high risk for misuse (e.g., pretrained language models, image generators, or scraped datasets)?
    \item[] Answer: \answerNA{} % Replace by \answerYes{}, \answerNo{}, or \answerNA{}.
    \item[] Justification: This research utilizes open-sourced datasets and language model weights.
    \item[] Guidelines:
    \begin{itemize}
        \item The answer NA means that the paper poses no such risks.
        \item Released models that have a high risk for misuse or dual-use should be released with necessary safeguards to allow for controlled use of the model, for example by requiring that users adhere to usage guidelines or restrictions to access the model or implementing safety filters. 
        \item Datasets that have been scraped from the Internet could pose safety risks. The authors should describe how they avoided releasing unsafe images.
        \item We recognize that providing effective safeguards is challenging, and many papers do not require this, but we encourage authors to take this into account and make a best faith effort.
    \end{itemize}

\item {\bf Licenses for existing assets}
    \item[] Question: Are the creators or original owners of assets (e.g., code, data, models), used in the paper, properly credited and are the license and terms of use explicitly mentioned and properly respected?
    \item[] Answer: \answerYes{} % Replace by \answerYes{}, \answerNo{}, or \answerNA{}.
    \item[] Justification: The using of existing datasets and foundation models is correctly cited. We obey the license and usage instructions.
    \item[] Guidelines:
    \begin{itemize}
        \item The answer NA means that the paper does not use existing assets.
        \item The authors should cite the original paper that produced the code package or dataset.
        \item The authors should state which version of the asset is used and, if possible, include a URL.
        \item The name of the license (e.g., CC-BY 4.0) should be included for each asset.
        \item For scraped data from a particular source (e.g., website), the copyright and terms of service of that source should be provided.
        \item If assets are released, the license, copyright information, and terms of use in the package should be provided. For popular datasets, \url{paperswithcode.com/datasets} has curated licenses for some datasets. Their licensing guide can help determine the license of a dataset.
        \item For existing datasets that are re-packaged, both the original license and the license of the derived asset (if it has changed) should be provided.
        \item If this information is not available online, the authors are encouraged to reach out to the asset's creators.
    \end{itemize}

\item {\bf New assets}
    \item[] Question: Are new assets introduced in the paper well documented and is the documentation provided alongside the assets?
    \item[] Answer: \answerYes{} % Replace by \answerYes{}, \answerNo{}, or \answerNA{}.
    \item[] Justification: We provide instructions to utilize our models for researching usage.
    \item[] Guidelines:
    \begin{itemize}
        \item The answer NA means that the paper does not release new assets.
        \item Researchers should communicate the details of the dataset/code/model as part of their submissions via structured templates. This includes details about training, license, limitations, etc. 
        \item The paper should discuss whether and how consent was obtained from people whose asset is used.
        \item At submission time, remember to anonymize your assets (if applicable). You can either create an anonymized URL or include an anonymized zip file.
    \end{itemize}

\item {\bf Crowdsourcing and research with human subjects}
    \item[] Question: For crowdsourcing experiments and research with human subjects, does the paper include the full text of instructions given to participants and screenshots, if applicable, as well as details about compensation (if any)? 
    \item[] Answer: \answerNA{} % Replace by \answerYes{}, \answerNo{}, or \answerNA{}.
    \item[] Justification: This research does not include crowdsourcing experiments or human subjects.
    \item[] Guidelines:
    \begin{itemize}
        \item The answer NA means that the paper does not involve crowdsourcing nor research with human subjects.
        \item Including this information in the supplemental material is fine, but if the main contribution of the paper involves human subjects, then as much detail as possible should be included in the main paper. 
        \item According to the NeurIPS Code of Ethics, workers involved in data collection, curation, or other labor should be paid at least the minimum wage in the country of the data collector. 
    \end{itemize}

\item {\bf Institutional review board (IRB) approvals or equivalent for research with human subjects}
    \item[] Question: Does the paper describe potential risks incurred by study participants, whether such risks were disclosed to the subjects, and whether Institutional Review Board (IRB) approvals (or an equivalent approval/review based on the requirements of your country or institution) were obtained?
    \item[] Answer: \answerNA{} % Replace by \answerYes{}, \answerNo{}, or \answerNA{}.
    \item[] Justification: This paper does not involve crowdsourcing nor research with human subjects.
    \item[] Guidelines:
    \begin{itemize}
        \item The answer NA means that the paper does not involve crowdsourcing nor research with human subjects.
        \item Depending on the country in which research is conducted, IRB approval (or equivalent) may be required for any human subjects research. If you obtained IRB approval, you should clearly state this in the paper. 
        \item We recognize that the procedures for this may vary significantly between institutions and locations, and we expect authors to adhere to the NeurIPS Code of Ethics and the guidelines for their institution. 
        \item For initial submissions, do not include any information that would break anonymity (if applicable), such as the institution conducting the review.
    \end{itemize}

\item {\bf Declaration of LLM usage}
    \item[] Question: Does the paper describe the usage of LLMs if it is an important, original, or non-standard component of the core methods in this research? Note that if the LLM is used only for writing, editing, or formatting purposes and does not impact the core methodology, scientific rigorousness, or originality of the research, declaration is not required.
    %this research? 
    \item[] Answer: \answerNA{} % Replace by \answerYes{}, \answerNo{}, or \answerNA{}.
    \item[] Justification: We illustrate the detail usage of LLM as an agent and give examples of prompts in the appendix to manage the robotic autonomy tasks in the pipeline. LLM does not impact the core methodology, scientific rigorousness, or originality of the research.
    \item[] Guidelines:
    \begin{itemize}
        \item The answer NA means that the core method development in this research does not involve LLMs as any important, original, or non-standard components.
        \item Please refer to our LLM policy (\url{https://neurips.cc/Conferences/2025/LLM}) for what should or should not be described.
    \end{itemize}

\end{enumerate}

\end{document}